\documentclass[sigconf,natbib=true,anonymous=false]{acmart}
\AtBeginDocument{%
  \providecommand\BibTeX{{%
    \normalfont B\kern-0.5em{\scshape i\kern-0.25em b}\kern-0.8em\TeX}}}

\usepackage{subfloat}
\usepackage{subfig}
\usepackage{diagbox}
\usepackage{multirow}

\begin{document}

\title{Is Summary Useful or Not?
\\ An Extrinsic Human Evaluation of Text Summaries on Downstream Tasks}

\author{Xiao Pu}
\email{puxiao@stu.pku.edu.cn}
\affiliation{%
  \institution{Peking University}
  \streetaddress{}
  \city{}
  \state{}
  \country{}
  \postcode{}
}

\author{Mingqi Gao}
\email{gaomingqi@pku.edu.cn}
\affiliation{%
  \institution{Peking University}
  \streetaddress{}
  \city{}
  \state{}
  \country{}
  \postcode{}
}

\author{Xiaojun Wan}
\email{wanxiaojun@pku.edu.cn}
\affiliation{%
  \institution{Peking University}
  \streetaddress{}
  \city{}
  \state{}
  \country{}
  \postcode{}
}

\renewcommand{\shortauthors}{}

\begin{abstract}
Research on automated text summarization relies heavily on human and automatic evaluation. While recent work on human evaluation mainly adopted intrinsic evaluation methods, judging the generic quality of text summaries, e.g. informativeness and coherence, our work focuses on evaluating the usefulness of text summaries with extrinsic methods. We carefully design three different downstream tasks for extrinsic human evaluation of summaries, i.e., question answering, text classification and text similarity assessment. We carry out experiments using system rankings and user behavior data to evaluate the performance of different summarization models. We find summaries are particularly useful in tasks that rely on an overall judgment of the text, while being less effective for question answering tasks. The results show that summaries generated by fine-tuned models lead to higher consistency in usefulness across all three tasks, as rankings of fine-tuned summarization systems are close across downstream tasks according to the proposed extrinsic metrics. Summaries generated by models in the zero-shot setting, however, are found to be biased towards the text classification and similarity assessment tasks, due to its general and less detailed summary style.

We further evaluate the correlation of 14 intrinsic automatic metrics with human criteria and show that intrinsic automatic metrics perform well in evaluating the usefulness of summaries in the question-answering task, but are less effective in the other two tasks. This highlights the limitations of relying solely on intrinsic automatic metrics in evaluating the performance and usefulness of summaries.



\end{abstract}




\maketitle

\section{Introduction}
Automated text summarization is a valuable tool as it allows one to quickly understand the main points of a longer document, by condensing the source text into a more concise version that still conveys the main points. The recent decade has witnessed the rapid development of automated text summarization models. One major challenge for the application of text summarization is how to evaluate whether such summaries generated by models are actually fluent, accurate, and useful.

Text summary evaluation methods can be divided into two categories: using automatic evaluation metrics or human judgments. Automatic evaluation metrics make it possible to evaluate the quality of generated text summaries in a much cheaper and quicker way, and existing popular automatic metrics are intrinsic evaluation metrics as they usually compare generated summaries with reference summaries or source documents to reflect the generic quality of summaries. Since current intrinsic automatic evaluation metrics can sometimes lead to erroneous conclusions \cite{he2022blind}, however, there is still no perfect substitute for human annotation. Human evaluation is usually used to evaluate the performance of text summarization models more reliably, or used as an oracle to evaluate the reliability of automated evaluation metrics.

There are two types of human evaluation: intrinsic evaluation and extrinsic evaluation. While intrinsic evaluation of text summarization focuses on the requirements of the task per se, e.g. coherence, fluency, and informativeness \cite{Fabbri2021,Bhandari2020}, extrinsic evaluation, also known as task-based evaluation, assesses the usefulness or helpfulness of text summaries in other tasks \cite{dorr2005methodology}. It is more objective and spontaneous because it evaluates human performance in a realistic usage scenario and is less demanding on the annotators. \cite{gillick2010non}

The prior works on extrinsic evaluation of summarization models have employed methods such as cross-comprehension tests \cite{kolluru2005subjectivity}, relevance judgment \cite{dorr2005methodology}, and question answering \cite{hirao2001extrinsic}. These studies are dated more than a decade ago. In recent years, neural summarization systems, especially those based on pre-trained language models have made great strides in intrinsic evaluation \cite{Fabbri2021,Bhandari2020}. However, to the best of our knowledge, no work has investigated the usefulness of these approaches from the perspective of extrinsic evaluation. Furthermore, these studies rely on a single method of extrinsic evaluation or are limited by the small scale of human experiments \cite{hovy1998}.
In light of these limitations, our work aims to propose a more comprehensive extrinsic evaluation method and conduct experiments on a larger scale, to systematically evaluate the usefulness of text summarization, including the summarization methods proposed recently. An attempt is also made to construct a trustworthy human-evaluated corpus, including subsets on three downstream tasks. 

Based on the proposed evaluation method in this study, we want to investigate the following research questions:
\begin{itemize}
    \item How useful are text summaries compared to the source articles?
    \item In which tasks are summaries more useful in general?
    \item What kind of summaries are more useful than others?
    \item Which intrinsic automatic metrics for text summarization correlate well with our human judgments?
\end{itemize}

The contributions of our work are summarized as follows: 
\begin{itemize}
    \item We introduce an extrinsic evaluation framework for systematically assessing the usefulness of text summarization. We also present seven extrinsic metrics in three downstream tasks.
    \item We annotate and construct a reliable human extrinsic evaluation dataset of 4,000 texts, including 400 source texts, 400 human summaries, and 3,200 summaries generated by eight different text summarization systems.
    \item We analyze the usefulness of various types of text summaries and discover that they are more useful in the classification task and the similarity assessment task.
    \item We re-evaluate 14 intrinsic automatic metrics through our proposed criteria and discover that most of them fail to reflect the extrinsic metrics in classification and similarity tasks.
\end{itemize}

The rest of this paper will be organized as follows: Section 2 introduces related work. Section 3 outlines the research methodology adopted in this study. Section 4 provides some preliminaries, including the datasets, summarization systems, and intrinsic automatic metrics utilized. The experimental setup is described in Section 5. The results of our analysis are presented in Section 6. Finally, significant conclusions are drawn in Section 7.

\section{Related Work}
\subsection{Intrinsic Evaluation for Summarization}
Past works that have assessed the quality of summaries through intrinsic evaluation methods can be classified into two main categories: intrinsic automatic metrics and intrinsic human evaluation. Early works evaluate summaries by computing the n-gram word overlap between reference summaries and generated summaries, such as BLEU \cite{bleu} and ROUGE \cite{lin2004rouge}, which have proven to be relatively effective over time. With the development of representation learning, researchers have proposed new intrinsic automatic metrics based on word embeddings, such as Greedy Matching \cite{greedymatching} and SMS \cite{clark2019sentence}, which compute the similarity of word embeddings between reference summaries and generated summaries. Additionally, automatic metrics based on question-answering \cite{summaqa} and entailment classification \cite{kryscinski-etal-2020-evaluating} have also been proposed. Human evaluation, on the other hand, is considered the gold standard for evaluating generated summaries. The Pyramid method \cite{Nenkova2004} serves as a viable framework for human evaluation, which has been further improved into a crowdsourcing method \cite{shapira2019crowdsourcing}.

Previous research has also investigated the relationship between intrinsic automatic metrics and intrinsic human judgments in the field of text summarization. A common approach to conduct meta-evaluation is to have annotators score the quality of summaries by Pyramid method \cite{Bhandari2020} or on multiple dimensions \cite{Fabbri2021} such as coherence, consistency, relevance, and fluency, and compute the correlation coefficient between the output scores of automatic evaluation metrics and human judgments. Prior research has shown significant differences in the performance of experts and non-experts in scoring summaries \cite{gillick2010non}. Recent work has examined the consistency between intrinsic automatic metrics and human preferences for different types of summaries and found that intrinsic automatic metrics cannot reliably evaluate summaries generated by models in the zero-shot setting. In contrast, our work investigates the correlation between intrinsic automatic metrics and extrinsic human judgments \cite{Goyal2022}.


\subsection{Extrinsic Evaluation for Summarization}
Previous work has acknowledged the human's subjectivity in evaluating summaries, and has attempted to alleviate this through the use of cross-comprehension tests \cite{kolluru2005subjectivity}. The usefulness of summaries has also been evaluated through a single extrinsic task, i.e. relevance judgment \cite{dorr2005methodology} and question answering \cite{hirao2001extrinsic}. While some researchers have proposed a set of tasks to measure the information content of full text and summaries, including a Shannon Game, a Question Game, and a Classification Game, finding that different extrinsic evaluation methods rate summaries differently, the scale of the experiments was too small to draw statistically significant conclusions \cite{hovy1998}.
Our work designs three distinct extrinsic evaluation tasks with a larger scale of human judgments and evaluates the summaries generated by the recently proposed summarization approaches.

\subsection{Summarization Models}
Summarization models can be broadly categorized into two groups: extractive and abstractive.
Extractive models directly identify and extract the most important sentences or words from the source text as the summary. Non-neural models, such as graph-based models, fuzzy logic-based models, and latent semantic analysis have been proposed and investigated \cite{mihalcea2004textrank,Lexrank,suanmali2009sentence,kyoomarsi2008optimizing,ozsoy2011text,mashechkin2011automatic}. Additionally, researchers have also explored extractive summarization based on neural network models \cite{nallapati2017summarunner,verma2017extractive,narayan2018ranking,liu2019fine}.
On the other hand, abstractive models generate a summary text that is not necessarily a direct extraction of the source text. In recent years, abstractive summarization models based on neural networks have been advancing and become dominant in the summarization field. A common paradigm is pre-training and fine-tuning \cite{liu-lapata-2019-text,BART,zhang2020pegasus}. Additionally, some prompt-based approaches have been proposed \cite{GPT3,T0}, enabling summarization models to learn from specific task instructions.

\section{Research Methodology}
The purpose of this study is to provide a comprehensive assessment of the usefulness of summaries in real-world usage scenarios. Participants are asked to complete three tasks using source articles and summaries, and their performance is measured to determine the usefulness of summaries.

\textbf{Measures of usefulness}
In our study, we consider a summary to be useful (or helpful) if it is able to facilitate users to complete a task. A useful summary should help users save time by being shorter than the source text, while also providing them with the important information they need to complete the task. 
Therefore, to assess the usefulness of the summaries, we decide to compare on two dimensions: time and correctness. Time refers to the amount of time it takes the participant to complete the task using either the source text or the summary. Correctness refers to the accuracy of the participant's response and is measured using different metrics for each task.  A web-based platform is developed and deployed for this study, to automatically record the completion time and submitted answers by participants for each task.

The three downstream tasks that we designed in this study are: \\\textbf{Question answering task}: In this task, participants are asked to answer questions based on the information provided in the source text or the summary. To evaluate the participant's accuracy, we use two commonly used evaluation metrics in QA systems to calculate the overlap between the answers submitted by the participant and the ground true answers. Additionally, we also propose a distinguished metric to reflect on the probability of the participants' answer attempts. By evaluating their performance in the QA task, we are able to determine the amount of useful information contained in the summary.

\textbf{Classification task}: In this task, participants are asked to select one or more tags based on the article or summary they see. The accuracy of their choices is calculated as a way of determining whether different types of summaries are useful in helping people make an overall judgment about the article.

\textbf{Similarity assessment task}: Participants are presented with a pair of news articles or summaries in this task. They are asked to take into account various factors such as the topic, event field, writing style, tone, etc. of the two articles to make a comprehensive judgment, and then score the similarity of the two articles or summaries on a scale of 1 to 4. By calculating how similar their scores are to the ground truth scores, we can determine how useful the summaries are for similarity judgments.

\section{Preliminaries}
\subsection{Datasets}
We use three datasets for different downstream tasks respectively in our study: 

CNN/DailyMail \cite{Hermann2015,Nallapati2016} is a widely used benchmark for text summarization, which includes a collection of news articles and their corresponding reference summaries that are typically 3-4 sentences in length. This dataset is used for extrinsic evaluation on the question answering task. 

New York Times Annotated Corpus \cite{NYT} contains a set of news articles along with human-written summaries. Each article is also associated with multiple tags or labels. This dataset is used for extrinsic evaluation on the text classification task.

The SemEval-2022 Task 8 dataset \cite{semeval} is a multilingual collection of the URLs of news articles that have been paired and annotated for their similarity level. The dataset includes nearly 1,000 article pairs from 18 different languages. This dataset is used for extrinsic evaluation on the text similarity assessment task.

\subsection{Representative Summarization Systems}
We need to select a few publicly available systems to generate summaries on the three datasets and then studying the usefulness of the summaries. As neural abstractive summarization methods with pretraining have achieved great success in recent years, we mainly focus on these summarization models. A total of six representative neural models are chosen as the abstractive systems, including:

\begin{itemize}
\item BART \cite{BART}: a sequence-to-sequence model trained as a denoising autoencoder, which is applicable to various natural language generation tasks. It is fine-tuned on CNN/DailyMail.
\item Pegasus \cite{zhang2020pegasus}: a model pre-trained with self-supervised gap-sentence-generation objective designed for abstractive summarization. We use the version fine-tuned on CNN/DailyMail.
\item BRIO \cite{liu2022BRIO}: a model with a new training paradigm that assigns candidate outputs probability mass according to their quality using contrastive learning. It is also fine-tuned on CNN/DailyMail.
\item T5 \cite{T5}: a text-to-text transfer learning framework that is pre-trained with several unsupervised and supervised objectives, including summarization. 
\item T0 \cite{T0}: a prompt-based model, which is fine-tuned on standard summarization datasets including CNN/DailyMail. 
\item GPT3 \cite{GPT3}: a prompt-based language model that achieves strong performance in the few-shot setting. In this work, we use OpenAI's text-davinci-002 \cite{ouyang2022}. 

\end{itemize}

We also include two simple extractive systems for comparison:
\begin{itemize}
\item Lead-n: Lead-3 is a simple but commonly used summarization baseline that selects the first three sentences of an article as the summary. we modify the Lead-3 setting and refer to it as the Lead-n model. Lead-n selects the first several sentences that are closest to the summary length we set.
\item Lexrank\cite{Lexrank}: a graph-based text summarization model that calculates the importance of sentences by determining the cosine similarity between them, and the sentences with highest scores are selected as the summary.  
\end{itemize}

\subsection{Intrinsic Automatic Metrics}
We employ a set of 14 automatic evaluation metrics to intrinsically assess the summaries. These metrics include n-gram overlap-based measures such as ROUGE-1, ROUGE-2, ROUGE-L \cite{lin2004rouge}, BLEU \cite{bleu}, METEOR \cite{meteor}, CIDEr \cite{cider} and CHRF \cite{chrf}. 
For metrics based on word embeddings, we report BERTScore \cite{zhang2019bertscore}, MoverScore \cite{moverscore}, Rouge-we \cite{rougewe}, Embedding average \cite{landauer1997solution},  Vector extrema \cite{forgues2014bootstrapping}, Greedy matching \cite{greedymatching}. Furthermore, we also include a model-based metric SummaQA \cite{summaqa} in our evaluation. All scores are reported in the range of 0-1. These scores will be compared with our extrinsic human evaluation results.

\section{Experimental Settings}
In this section, we present the construction and annotation of three datasets for use and the design of our user study for extrinsic evaluation. Specifically, we focus on three downstream tasks: question answering (QA), text classification, and text similarity assessment. We then propose extrinsic metrics based on these tasks.

\subsection{Data Preparation Process}
\textbf{Processing and annotating datasets.}
We reprocess and manually annotate three existing datasets for use in our user study. The datasets for the downstream tasks are constructed in the following steps.

For the QA task, we randomly select 100 pairs of source text and reference summary from the CNN/DailyMail test set. We then construct two datasets for the QA task, namely QA-ref and QA-source. For QA-ref, we formulate four questions and their corresponding answers for each reference summary. For QA-source, we read the longer source text, identify the important points within the news, and formulate four questions accordingly. For each question, we search for all corresponding content within the source text as the correct answer. In both datasets, a question may have multiple correct answers.

For the classification task, we randomly sample 100 news articles from the New York Times Annotated Corpus test set and obtain 19 tags. We analyze these tags and identify those that are vague in meaning and difficult to identify from the article, such as 'Front Page', and those that were redundant and dependent on other tags, such as 'Travel', 'Theater', 'Dining and Wine', and 'Movies', which always appear alongside 'Art'. We remove these tags and retain a total of 11 tags, at least one for each news article.
  
For the similarity task, we utilize the Semeval2022 task8 dataset and construct a dataset for use consisting of 100 pairs of news articles, together with reference summaries and corresponding similarity scores through the following steps: First, we randomly crawl 300 pairs of news pages through the corresponding URLs and extract the title, description, and body parts of each article. The next step is data cleaning, where we remove pairs with empty or too short titles/descriptions/bodies and those whose descriptions are directly sourced from the beginning of the source article with incomplete sentences. Then, we splice headlines and descriptions to form summaries. After manual review, we finally retain 100 pairs of news articles for the similarity task, comprising 200 news articles.

\textbf{Generating summaries of similar length.}
In order to eliminate any potential bias that may have resulted from variations in text length, we keep the length of summaries within a defined range. This range is determined based on the average length of the human summaries in each task. To achieve this, we employ a two-step process:  First, we set a range for the number of tokens generated for abstractive models and a range for the number of sentences generated for extractive models, during the process of generating the summaries with the model. Secondly, all summaries are truncated to the established range. Figure \ref{tab:sum_len} in the appendix shows the length of summaries in the three tasks.

\subsection{Web-based Platform for Experiments}
\begin{figure}[h]
\centering
\includegraphics[width=\linewidth]{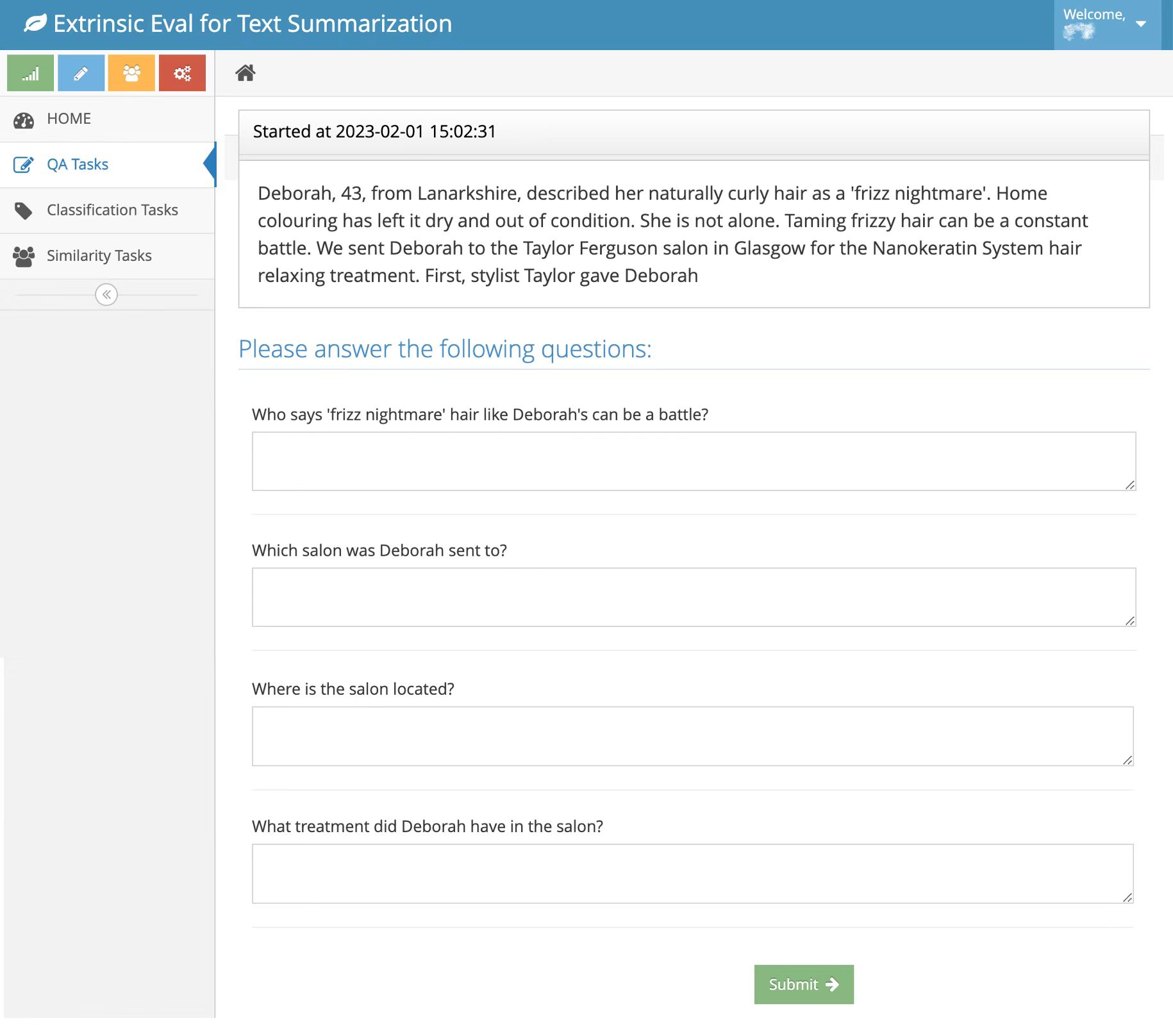}
\caption{A screenshot of the answer page for the QA task. The user information on the platform has been anonymized.}
\label{fig:platform}
\end{figure}

We implement a web-based platform (as shown in Figure \ref{fig:platform}) to facilitate users' participation in the tasks and the acquisition of experiment data, which includes responses and completion time for each question. To guarantee impartiality, the platform is designed to prohibit the utilization of the copy-paste/search functionality. Furthermore, the website offers guidance information and exemplar answers to assist participants to fully understand the tasks.

\subsection{Experimental Details}
We initially recruit ten individuals to participate in the QA-ref, classification and similarity tasks. For the QA-source task, we conduct a separate recruitment process and select another ten individuals. The purpose of this design was to ensure that participants had no prior memory of the text or question content. By having different individuals perform each task, we aim to minimize the influence of previously seen summaries on their responses to the original text questions. In total, we collect 1,000 responses for each task, resulting in a dataset of 10,000 annotations.

To maintain the quality of annotations, all participants are recruited from the university campus, they are all graduate students aged between 22 and 26. All participants had the same native language and are proficient in English as their second language. They have obtained excellent scores in internationally recognized English exams, indicating their suitability for successfully completing the experimental tasks.

To ensure that the participants' responses are only based on the content of the text currently being viewed and to minimize the influence of individual differences, a method for distributing the texts is devised. The following considerations are taken into account:
1)  To prevent people from having an advantage due to prior exposure to a similar text, each person is allowed to see only one text (either source text or summary) from the same source.
2)  To ensure fairness and remove the influence of individual differences, each person must be exposed to the same number of texts from each system, regardless of their proficiency level.

The distribution method is as follows: One source text is associated with nine summaries (including reference summary), resulting in ten texts (including source text) originating from the same source text. 

First, all summaries are aligned with the source text, then different systems are arranged in the following order: [Source, Human, BART, Pegasus, Lexrank, Lead-n, BRIO, T5, T0, GPT3]. After that, all texts are numbered, with text\_id (0-999) as their unique identifier. Therefore, the hundreds place indicates the system corresponding to the text, and the tens place and the individual place indicate the corresponding source text.

The texts are assigned to different participants according to the system it belongs to and the corresponding source text. Each participant is assigned to a user\_id and the correspondence between texts and participants is established by the following formula:
$$
y = \lfloor \frac{text\_id-\lfloor \frac{text\_id}{100}\rfloor \times 100}{10}\rfloor - \lfloor \frac{text\_id}{100}\rfloor 
$$

$$
user\_id (y)=
\begin{cases} 
y,&y \geq 0
\\
10+y,&y < 0
\\ \end{cases}
$$
\subsection{Proposed Extrinsic Metrics}
Based on the three downstream tasks, we propose the following extrinsic metrics to evaluate the usefulness of the summaries.

For the QA task, let $y_n^k$denote the participant's answer to the $\it{k}$-th question of $\it{n}$-th article. All the correct answers to a question are ordered and  $\hat{y}_n^{ki}$ denote the $\it{i}$-th key answer to the $\it{k}$-th question of $\it{n}$-th article. $N$ represents the number of summaries of each system, which equals 100, and $K$ represents the number of questions for each article, which equals 4 in this case, and the three metrics are calculated as follows. 
\begin{itemize}
\item Answerable measures the proportion of questions that can be answered according to the text.
\item Exact Match Ratio (EM), which counts the overall accuracy rate of the answers. EM of each system is calculated as:
$$
EM = \frac{1}{NK}\sum_{n=1}^{N}\sum_{k=1}^{K}MAX_i(I(y_n^k==\hat{y}_n^{ki}))
$$
$$ with \ \
I(y_n^k==\hat{y_n}^{ki})=
\begin{cases} 
1,& y_n^k=\hat{y_n}^{ki}
\\
0,& y_n^k\neq\hat{y_n}^{ki}
\\ \end{cases}
$$

\item F1 is a looser measure of the average overlap between the prediction and ground truth answer. When calculating F1, both $y_n^k$ and $\hat{y_n}^k$ are tokenized into sets of words. F1 is calculated as  
$$F_1 =  \frac{1}{NK}\sum_{n=1}^{N}\sum_{k=1}^{K} MAX_i\frac{2\lvert y_n^{k}\cap \hat{y_n}^{ki} \rvert}{\lvert y_n^{k} \rvert + \lvert\hat{y_n}^{ki}\rvert}
$$
\end{itemize}
\begin{table*}[h]
\resizebox{\textwidth}{!}{%
\begin{tabular}{@{}lcccccccccccccc@{}}
\toprule
\multicolumn{1}{c}{\multirow{2}{*}{system}} &
  \multicolumn{4}{c}{QA (ref-based)} &
  \multicolumn{4}{c}{QA (source-based)} &
  \multicolumn{3}{c}{Classification} &
  \multicolumn{3}{c}{Similarity} \\ \cmidrule(l){2-15} 
\multicolumn{1}{c}{} &
  answerable &
  EM &
  F1 &
  time(seconds) &
  answerable &
  EM &
  F1 &
  time(seconds) &
  EM &
  F1 &
  time(seconds) &
  MSE &
  $\rho$ &
  time(seconds) \\ \midrule
source &
  0.8550 &
  0.3225 &
  0.5077 &
  280.04 &
  \textbf{0.8875} &
  \textbf{0.5050} &
  \textbf{0.6796} &
  211.64 &
  0.8827 &
  0.8951 &
  72.97 &
  0.9136 &
  0.6184 &
  37.74 \\
reference &
  \textbf{0.8875} &
  \textbf{0.5400} &
  \textbf{0.7535} &
  \textbf{93.94} &
  0.5375 &
  0.2725 &
  0.3746 &
  83.3 &
  \textbf{0.9127} &
  \textbf{0.9156} &
  34.37 &
  0.7736 &
  0.7060 &
  \textbf{19.92} \\
bart &
  0.4975 &
  0.2400 &
  0.3240 &
  108.37 &
  0.4900 &
  0.2325 &
  0.3197 &
  \textbf{83.05} &
  0.8964 &
  0.9015 &
  \textbf{25.43} &
  0.9803 &
  0.6085 &
  21.94 \\
pegasus &
  0.5475 &
  0.2100 &
  0.3222 &
  112.55 &
  0.5125 &
  0.2825 &
  0.3662 &
  89.66 &
  0.8900 &
  0.8942 &
  29.88 &
  0.9836 &
  0.6014 &
  23.93 \\
lexrank &
  0.3625 &
  0.0900 &
  0.1631 &
  111.78 &
  0.3775 &
  0.1500 &
  0.2291 &
  92.01 &
  0.9000 &
  0.9017 &
  29.88 &
  1.2403 &
  0.5323 &
  23.77 \\
Lead-n &
  0.4175 &
  0.1600 &
  0.2483 &
  110.78 &
  0.4775 &
  0.2475 &
  0.3342 &
  84.33 &
  0.8773 &
  0.8792 &
  31.29 &
  1.4336 &
  0.4536 &
  23.42 \\
BRIO &
  0.5825 &
  0.2350 &
  0.3598 &
  104.21 &
  0.5425 &
  0.3075 &
  0.4040 &
  90.15 &
  0.9000 &
  0.9036 &
  25.9 &
  \textbf{0.7569} &
  0.6998 &
  21.07 \\
t5 &
  0.4400 &
  0.1600 &
  0.2416 &
  106.07 &
  0.4375 &
  0.2075 &
  0.2861 &
  86.57 &
  0.8791 &
  0.8814 &
  34.86 &
  1.3736 &
  0.4699 &
  20.17 \\
t0 &
  0.5350 &
  0.1875 &
  0.3003 &
  107.21 &
  0.5100 &
  0.2600 &
  0.3530 &
  98.6 &
  0.8864 &
  0.8889 &
  28.57 &
  0.7669 &
  \textbf{0.7087} &
  20.96 \\
gpt3 &
  0.4200 &
  0.1575 &
  0.2338 &
  100.02 &
  0.4500 &
  0.1975 &
  0.2855 &
  83.74 &
  0.9036 &
  0.9068 &
  29.11 &
  0.8469 &
  0.6741 &
  20.66 \\ \bottomrule
\end{tabular}%
}
\caption{Usefulness of different systems on downstream tasks, including the average time taken by participants to complete tasks with different system outputs and results of extrinsic metrics based on user performance.}
\label{tab:sys_exmetrics}
\end{table*}

For the classification task, we use EM and F1, two metrics that are commonly used in multiclass classification tasks.

For the similarity task, we use the following metrics:
\begin{itemize}
    \item Mean Squared Error (MSE), which indicates the extent to which the participant's answer deviates from the standard answer.
\item Spearman's $\rho$, a measure of the correlation between the participant's judgment and the true similarity. It can only be used for system-level analysis because it cannot be calculated using separate texts.
\end{itemize}

\section{Results and analysis}

\subsection{Analyzing Our Extrinsic Metrics}
\begin{figure}[h]
\centering
\includegraphics[width=0.7\linewidth]{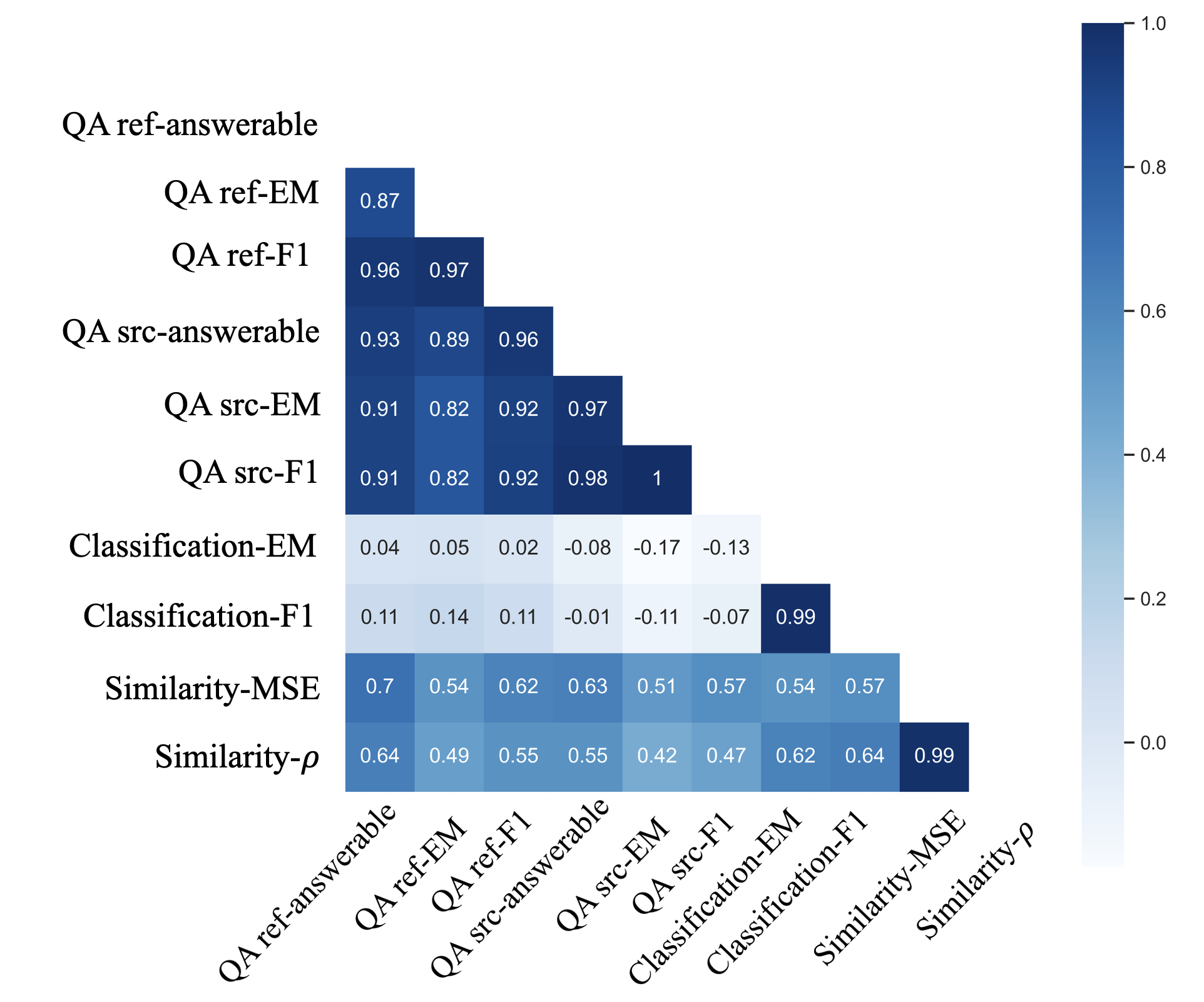}
\caption{System-level Pearson correlation of all extrinsic metrics. The result of Kendall correlation is shown in Figure \ref{tab:ex-sysl-kendall} in the Appendix.}
\label{fig:syscorrelation}
\end{figure}

In this section, we study the relationship between our proposed extrinsic metrics. We compute system-level correlations of all the extrinsic metrics (as shown in Figure \ref{fig:syscorrelation}). 

According to the Pearson's \textit{r}, extrinsic metrics of the same downstream task are highly correlated, ranging from 0.8 to 1. QA-ref and QA-source are highly correlated at system level, with Pearson's \textit{r} above 0.8 and Kendall's $\tau$ above 0.69. This suggests that there is little difference in the relative performance of the systems on QA-ref and QA-source, although they differ in the way the dataset is constructed. 
Comparing the metrics of the different downstream tasks, we find that the QA task and the classification task are poorly correlated, with Pearson's \textit{r} ranging from -0.2 to 0.2. Whereas the similarity task is moderately correlated with both the other two tasks, with Pearson's \textit{r} ranging from 0.4 to 0.7.  
Overall, moderate to weak correlations illustrate that our experiment involves three tasks of different perspectives to measure the usefulness of the summary.

\subsection{Evaluating Usefulness of Summaries}
In this section, we compare the performance of different summarization systems by means of the proposed extrinsic evaluation method (as shown in Table \ref{tab:sys_exmetrics}) and try to answer some questions regarding the usefulness of summaries.

\begin{table*}[]
\resizebox{\textwidth}{!}{%
\begin{tabular}{@{}lllllllllllllllll@{}}
\toprule
 &
  \multicolumn{8}{c}{QA (ref-based)} &
  \multicolumn{8}{c}{QA (source-based)} \\ \cmidrule(l){2-17} 
\multirow{-2}{*}{} &
  \multicolumn{2}{c}{Answerable} &
  \multicolumn{2}{c}{EM} &
  \multicolumn{2}{c}{F1} &
  \multicolumn{2}{c}{Time(seconds)} &
  \multicolumn{2}{c}{Answerable} &
  \multicolumn{2}{c}{EM} &
  \multicolumn{2}{c}{F1} &
  \multicolumn{2}{c}{Time(seconds)} \\ \midrule
Source &
  0.86 &
   &
  0.32 &
   &
  0.51 &
   &
  280 &
   &
  0.89 &
   &
  0.51 &
   &
  0.7 &
   &
  212 &
   \\
Reference Summaries &
  0.89 &
  {\color[HTML]{FE0000} +4\%} &
  0.54 &
  {\color[HTML]{FE0000} +67\%} &
  0.75 &
  {\color[HTML]{FE0000} +48\%} &
  94 &
  {\color[HTML]{FE0000} -66\%} &
  0.54 &
  {\color[HTML]{009901} -39\%} &
  0.27 &
  {\color[HTML]{009901} -46\%} &
  0.4 &
  {\color[HTML]{009901} -45\%} &
  88 &
  {\color[HTML]{FE0000} -58\%} \\
All Summaries &
  0.52 &
  {\color[HTML]{009901} -39\%} &
  0.22 &
  {\color[HTML]{009901} -32\%} &
  0.33 &
  {\color[HTML]{009901} -36\%} &
  106 &
  {\color[HTML]{FE0000} -62\%} &
  0.52 &
  {\color[HTML]{009901} -41\%} &
  0.24 &
  {\color[HTML]{009901} -53\%} &
  0.3 &
  {\color[HTML]{009901} -52\%} &
  83 &
  {\color[HTML]{FE0000} -61\%} \\ \bottomrule
\end{tabular}%
}
\caption{Summaries compared to source texts in the QA tasks. The red percentages indicate that summaries are better compared to the source text, i.e. participants take less time or perform better in completing the task. The green ones indicate the opposite. Although the summaries represent a significant time saving, participants perform worse in QA tasks using the summaries compared to source texts.}
\label{tab:src_sum_1}
\end{table*}

\begin{table*}[]
\resizebox{0.9\textwidth}{!}{%
\begin{tabular}{lcccccclcccccc}
\hline
       & \multicolumn{6}{c}{Classification} &  & \multicolumn{6}{c}{Similarity} \\ \cline{2-14} 
\multirow{-2}{*}{} &
  \multicolumn{2}{c}{EM} &
  \multicolumn{2}{c}{F1} &
  \multicolumn{2}{c}{Time(seconds)} &
   &
  \multicolumn{2}{c}{MSE} &
  \multicolumn{2}{c}{Spearman's $\rho$} &
  \multicolumn{2}{c}{Time(seconds)} \\ \hline
Source & 0.88   &    & 0.90   &   & 73  &   &  & 0.91  &   & 0.6  &   & 38  &   \\
Reference Summaries &
  0.91 &
  {\color[HTML]{FE0000} +3\%} &
  0.92 &
  {\color[HTML]{FE0000} +2\%} &
  34 &
  {\color[HTML]{FE0000} -53\%} &
   &
  0.77 &
  {\color[HTML]{FE0000} -15\%} &
  0.7 &
  {\color[HTML]{FE0000} +14\%} &
  20 &
  {\color[HTML]{FE0000} -47\%} \\
All Summaries &
  0.89 &
  {\color[HTML]{FE0000} +1\%} &
  0.90 &
  - &
  30 &
  {\color[HTML]{FE0000} -59\%} &
   &
  1.02 &
  {\color[HTML]{009901} +11\%} &
  0.6 &
  - &
  22 &
  {\color[HTML]{FE0000} -42\%} \\ \hline
\end{tabular}%
}
\caption{Summaries compared to source texts in the classification and similarity tasks. It shows that summary serves about the same function as the source text in these two tasks, and even helps participants to do tasks better.}
\label{tab:src_sum_2}
\end{table*}

\textbf{How useful are text summaries compared to source articles?}

Results from three downstream tasks demonstrate that the use of summaries significantly reduces the time required for task completion. Specifically, compared to the source articles, the average time participants spent using summaries to complete QA tasks drops by 61-62\% (as shown in Table \ref{tab:src_sum_1}). Similar results can also be observed in the classification and similarity tasks, with the time-saving percentages of 59\% and 42\%, respectively (as shown in Table \ref{tab:src_sum_2}).

We also find that \textbf{summaries are particularly useful in classification and similarity tasks}. In the QA task, source texts outperform summaries on average, while in the classification and similarity tasks, participants spend less time as well as perform better with summaries. This may be due to the fact that making an overall judgment about the text, such as classification or similarity assessment, does not require as much information as answering specific questions. As a result, the excess information in the long source text may not aid in decision-making and even interfere with human judgments. This is supported by observed people's tendencies in the classification task, where they tend to assign more tags to longer source articles, potentially leading to a higher recall but lower precision in comparison to the human summaries.

\textbf{Difference between QA-ref and QA-source} In the QA-source task, where questions and answers are constructed from the source text, source articles excel in all three metrics (answerable, EM, and F1). In the QA-ref task, where questions and answers are constructed from the reference summary, although the answerable metric is similar for source articles and reference summaries, in terms of the other two metrics, i.e. EM and F1, reference summaries are approximately 50\% better than the source text.
This is because the information in the reference summary is only a subset of the source text. Therefore in some cases, although people find some questions in QA-ref answerable by looking at the source text, their answers may be counted as incorrect because they do not appear in the reference summary (even though they may be correct according to the source text). 

\begin{figure}[h]
\centering
\includegraphics[width=0.8\linewidth]{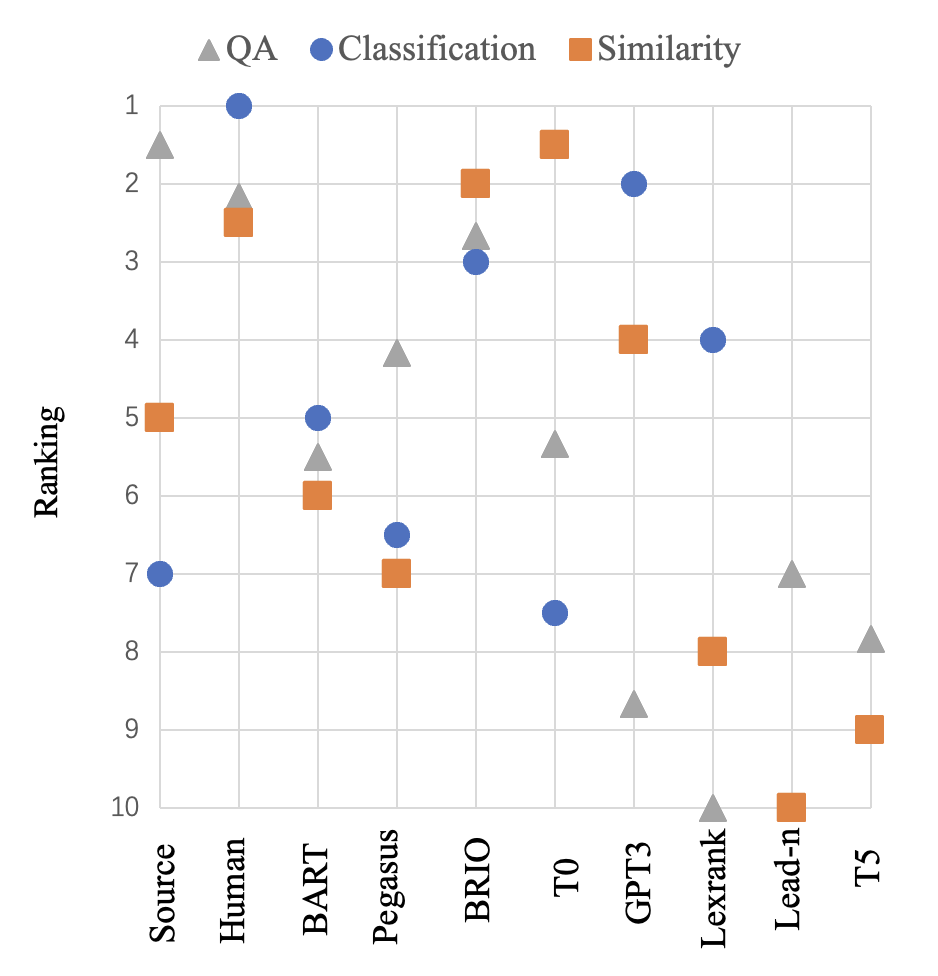}
\caption{Average ranking of different systems on three different tasks. Each ranking is calculated by averaging the rankings over extrinsic metrics for the same task.}
\label{tab:sys_rank}
\end{figure}

\textbf{What kind of summaries are more useful?}
We divide all the automated summaries into three categories based on the model used to generate them:  fine-tuned,  prompt-based, and simple extractive. A question we want to know is, how stable or consistent is the usefulness level of summaries across different downstream tasks? By analyzing rankings of the source text and summaries in the three tasks, as is shown in Figure \ref{tab:sys_rank}, we find that:
\textbf{The summaries generated by fine-tuned models have higher consistency in usefulness across different tasks}, such as those generated by BART, Pegasus, and BRIO, with a stable ranking similar to that of the human summaries. This suggests that summaries generated by fine-tuned models are insensitive to differences between tasks.\textbf{The summaries generated by simple extractive models and models in the zero-shot setting exhibit a varying ranking across tasks.} For example, both zero-shot GPT3 summaries and simple extractive Lexrank summaries show high or above average rankings in the classification task, medium rankings in the similarity task, and very low rankings in the QA task.

\begin{table}[]
\resizebox{\columnwidth}{!}{%
\begin{tabular}{|l|}
\hline
\textbf{Source text:} \\
\begin{tabular}[c]{@{}l@{}}A heartbroken pensioner is believed to have killed himself six days \\ after his wife 's death by jumping from a bridge at their ' special \\ place ' where they used to take romantic walks together. {[}...{]} Today \\ officers confirmed a body pulled from the River Trent on April 15 \\ by a specialist underwater search unit was sadly that of the missing \\ pensioner. {[}...{]} June tragically died on March 31 , eight hours after \\ collapsing suddenly from what doctors at the Queen 's Medical \\ Centre in Nottingham described as a ' catastrophic bleed ' to the \\ brain. {[}...{]}\end{tabular} \\ \\
\textbf{GPT3 summary:} \\
\begin{tabular}[c]{@{}l@{}}A man is believed to have killed himself by jumping from a bridge \\ at a picturesque spot where he and his wife used to take romantic \\ walks together, six days after she died from a brain hemorrhage.\end{tabular} \\ \\
\textbf{BRIO summary:} \\
\begin{tabular}[c]{@{}l@{}}John Lord , 86 , went missing from his home on April 6 less than a \\ week after his beloved wife June , 81, died from a ' catastrophic \\ bleed ' to the brain. The body of the pensioner was recovered from \\ the River Trent on April 15. His family believe he may have jumped \\ from a bridge at the picturesque beauty {[}...{]}\end{tabular} \\ \\
\textbf{T0 summary:} \\
\begin{tabular}[c]{@{}l@{}}John Lord, 86, went missing from his home on April 6. His wife \\ June, 81, died from a 'catastrophic bleed' to the brain. Family feared \\ the worst after finding a note describing how much he missed her. \\ Mr Lord's body was pulled from the River Trent on April 15.\end{tabular} \\ \\
\textbf{Lexrank summary:} \\
\begin{tabular}[c]{@{}l@{}}Mr Lord , 86 , went missing from his home in St Ann 's on Monday,\\ 6 April . ' A Nottinghamshire Police spokesperson said : ' The body \\ of a man found in the River Trent on April 15 , 2015 , has been \\ confirmed as that of missing John Lord . ' Message :Mr Lord 's \\ daughter Alison said her father was grieving and had left a heart-\\ breaking note signed {[}...{]}\end{tabular} \\ \hline
\end{tabular}%
}
\caption{A case study to illustrate the difference of summary style. By looking at the source text and the summaries generated with different models, we find that the zero-shot GPT3 summary tends to paraphrase the news in a more general way, making it easier for readers to capture the main point, but often omitting detailed information. Instead, summaries of fine-tuned BRIO and T0 models contain more detailed information, making it more suitable for QA tasks. The coherence between sentences in the extractive Lexrank summary is poor, causing difficulty in reading.}
\label{tab:casestudy}
\end{table}

\begin{table*}[tbh]
\resizebox{\textwidth}{!}{%
\begin{tabular}{p{30mm}|cccccccccccccccccccc}
\hline
\multirow{3}{*}{\diagbox[width=33mm,height=11mm]{Automatic Metrics}{Extrinsic Criteria}} &
  \multicolumn{6}{c}{QA (ref-based)} &
  \multicolumn{6}{c}{QA (source-based)} &
  \multicolumn{4}{c}{Classification} &
  \multicolumn{4}{c}{Similarity} \\ \cline{2-21} 
 &
  \multicolumn{2}{c}{answerable} &
  \multicolumn{2}{c}{EM} &
  \multicolumn{2}{c}{F1} &
  \multicolumn{2}{c}{answerable} &
  \multicolumn{2}{c}{EM} &
  \multicolumn{2}{c}{F1} &
  \multicolumn{2}{c}{EM} &
  \multicolumn{2}{c}{F1} &
  \multicolumn{2}{c}{MSE} &
  \multicolumn{2}{c}{$\rho$} \\ \cline{2-21} 
 &
  \textit{r} &
  $\tau$ &
  \textit{r} &
  $\tau$ &
  \textit{r} &
  $\tau$ &
  \textit{r} &
  $\tau$ &
  \textit{r} &
  $\tau$ &
  \textit{r} &
  $\tau$ &
  \textit{r} &
  $\tau$ &
  \textit{r} &
  $\tau$ &
  \multicolumn{1}{c}{\textit{r}} &
  $\tau$ &
  \textit{r} &
  $\tau$ \\ \hline
ROUGE-1 &
  0.95** &
  0.71* &
  0.94** &
  0.76** &
  0.98** &
  0.86** &
  0.95** &
  0.79** &
  0.89** &
  0.64* &
  0.91** &
  0.64* &
  0.51 &
  0.50 &
  0.56 &
  0.50 &
  0.48 &
  0.43 &
  0.40 &
  0.36 \\
ROUGE-2 &
  0.97** &
  0.79** &
  0.94** &
  0.91** &
  0.98** &
  0.93** &
  0.92** &
  0.71* &
  0.89** &
  0.71* &
  0.89** &
  0.71* &
  0.23 &
  0.21 &
  0.29 &
  0.21 &
  0.18 &
  0.29 &
  0.10 &
  0.36 \\
ROUGE-L &
  0.99** &
  0.93** &
  0.93** &
  0.76** &
  0.97** &
  0.79** &
  0.91** &
  0.71* &
  0.87** &
  0.71* &
  0.87** &
  0.71* &
  0.33 &
  0.43 &
  0.40 &
  0.43 &
  0.29 &
  0.29 &
  0.22 &
  0.36 \\
BLEU &
  0.89** &
  0.64* &
  0.88** &
  0.84** &
  0.92** &
  0.93** &
  0.85** &
  0.71* &
  0.83* &
  0.71* &
  0.83* &
  0.71* &
  0.21 &
  0.21 &
  0.28 &
  0.21 &
  -0.01 &
  0.14 &
  -0.08 &
  0.21 \\
METEOR &
  0.93** &
  0.64* &
  0.88** &
  0.84** &
  0.94** &
  0.79** &
  0.91** &
  0.86** &
  0.87** &
  0.71* &
  0.89** &
  0.71* &
  0.49 &
  0.50 &
  0.54 &
  0.50 &
  0.31 &
  0.36 &
  0.24 &
  0.29 \\
CHRF &
  0.95** &
  0.64* &
  0.90** &
  0.84** &
  0.96** &
  0.93** &
  0.91** &
  0.71* &
  0.88** &
  0.71* &
  0.89** &
  0.71* &
  0.48 &
  0.50 &
  0.52 &
  0.50 &
  0.31 &
  0.29 &
  0.23 &
  0.36 \\
CIDEe &
  0.75* &
  0.50 &
  0.83** &
  0.69* &
  0.85** &
  0.79** &
  0.82* &
  0.71* &
  0.82* &
  0.57 &
  0.83* &
  0.57 &
  0.12 &
  0.00 &
  0.20 &
  0.00 &
  -0.03 &
  0.07 &
  -0.09 &
  0.00 \\
BERTScore &
  0.94** &
  0.71* &
  0.87** &
  0.62* &
  0.93** &
  0.71* &
  0.89** &
  0.93** &
  0.85** &
  0.79** &
  0.86** &
  0.79** &
  0.54 &
  0.43 &
  0.59 &
  0.43 &
  0.54 &
  0.43 &
  0.48 &
  0.36 \\
MOVERScore &
  0.97** &
  0.79** &
  0.93** &
  0.69* &
  0.97** &
  0.79** &
  0.93** &
  0.86** &
  0.87** &
  0.71* &
  0.88** &
  0.71* &
  0.55 &
  0.50 &
  0.60 &
  0.50 &
  0.46 &
  0.43 &
  0.39 &
  0.36 \\
ROUGE-we &
  0.95** &
  0.71* &
  0.94** &
  0.76** &
  0.98** &
  0.86** &
  0.95** &
  0.79** &
  0.90** &
  0.64* &
  0.91** &
  0.64* &
  0.50 &
  0.50 &
  0.55 &
  0.50 &
  0.45 &
  0.43 &
  0.38 &
  0.36 \\
EmbeddingAverage &
  0.79* &
  0.50 &
  0.82* &
  0.69* &
  0.86** &
  0.79** &
  0.87** &
  0.71* &
  0.85** &
  0.57 &
  0.86** &
  0.57 &
  0.71* &
  0.57 &
  0.75 &
  0.57 &
  0.56 &
  0.50 &
  0.51 &
  0.43 \\
VectorExtrema &
  0.80* &
  0.57 &
  0.80* &
  0.76** &
  0.86** &
  0.86** &
  0.82* &
  0.64* &
  0.84** &
  0.64* &
  0.84** &
  0.64* &
  0.37 &
  0.21 &
  0.42 &
  0.21 &
  0.40 &
  0.36 &
  0.33 &
  0.29 \\
GreedyMatching &
  0.89** &
  0.64* &
  0.80* &
  0.69* &
  0.88** &
  0.79** &
  0.85** &
  0.71* &
  0.85** &
  0.71* &
  0.86** &
  0.71* &
  0.60 &
  0.50 &
  0.64 &
  0.50 &
  0.43 &
  0.50 &
  0.36 &
  0.43 \\
SummaQA &
  0.87** &
  0.57 &
  0.85** &
  0.62* &
  0.91** &
  0.71* &
  0.93** &
  0.79** &
  0.87** &
  0.64* &
  0.89** &
  0.64* &
  0.24 &
  0.21 &
  0.30 &
  0.21 &
  0.43 &
  0.43 &
  0.35 &
  0.36 \\ \hline
\end{tabular}%
}
\caption{Pearson's \textit{r} and Kendall's $\tau$ between intrinsic automatic metrics and extrinsic Criteria. Significance is indicated by * for p-values less than or equal to 0.05 and ** for p-values less than or equal to 0.01.}
\label{tab:sysl_auto}
\end{table*}

We also identify differences in the style of the summaries generated by the different models and a case study in Table \ref{tab:casestudy} illustrates this point. \textbf{The Summaries generated by fine-tuned models tend to be more informative and specific}, including more factual details such as times, places, and numbers. \footnote{It's important to note that this observation is only based on the summaries fine-tuned on the CNN/DailyMail dataset. Fine-tuning on other datasets may produce different results and therefore cannot be generalized as all summaries generated by fine-tuned models. When referring to summaries generated by fine-tuned models, it should only be understood as those fine-tuned on the CNN/DailyMail dataset.}Due to this trait, summaries generated by fine-tuned models are found to be more useful for detail-oriented QA tasks, compared to their counterparts. The top six in all systems except the source text and reference summary are fine-tuned models, including task-specific fine-tuned T0 system. \textbf{Summaries generated by models in the zero-shot setting are more abstractive and general} than that of fine-tuned models, and therefore they are found to be more suitable for tasks that require overall judgment, such as classification and similarity tasks. As is shown in Figure \ref{tab:sys_rank}, zero-shot GPT3 summaries rank second in the  classification task but only second to last in the QA task.

Compared to them, simple extractive summaries are more coarse-grained and less useful. According to the case study, they contain relatively less important information in a limited space. These two models were developed in the early years of natural language processing, and after nearly two decades of advancements in the field, their usefulness has been surpassed by more recent models.

\subsection{Evaluating Intrinsic Automatic Metrics}
\begin{figure}[h]
\centering
\includegraphics[width=\linewidth]{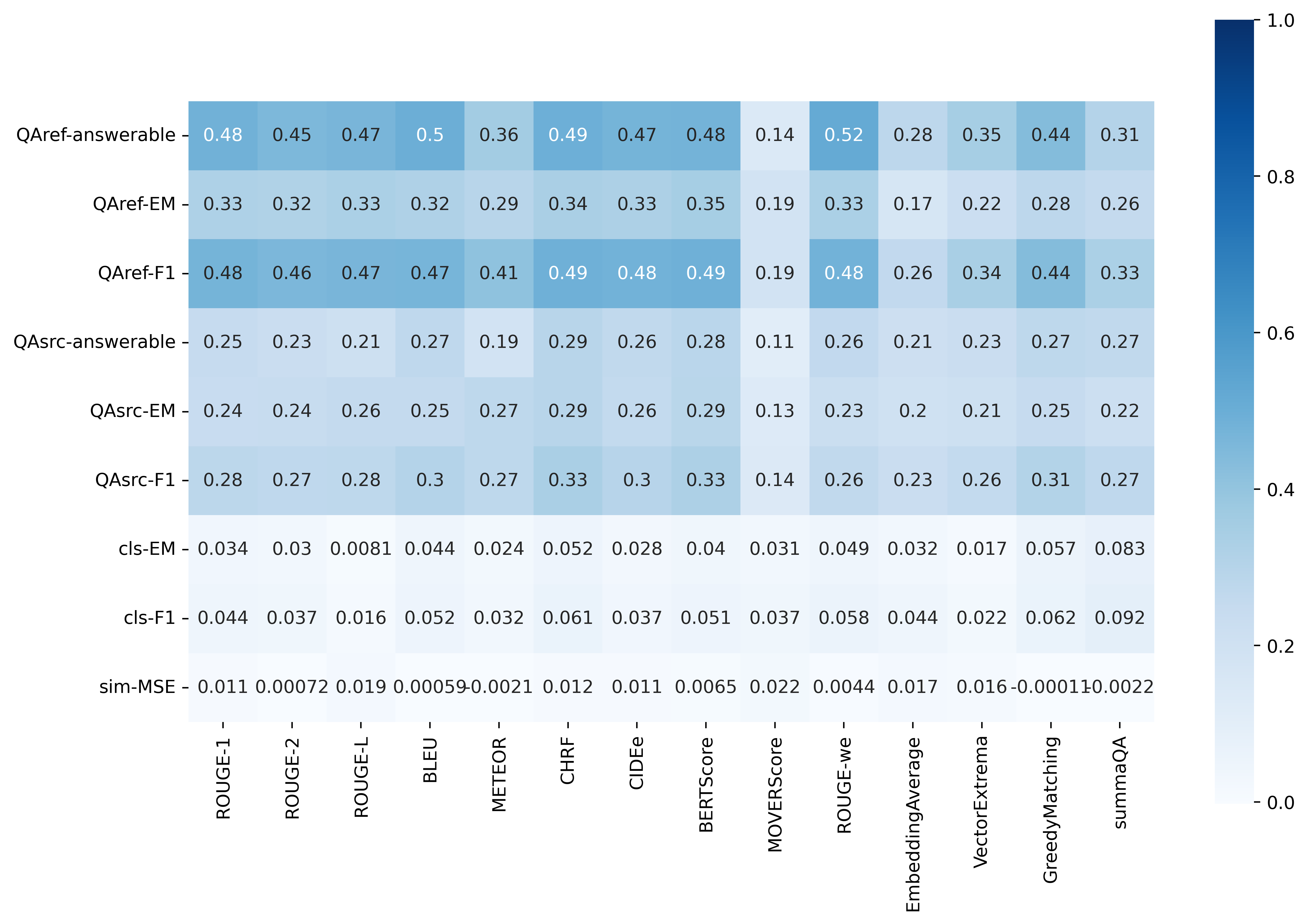}
\caption{Summary-level correlation between intrinsic automatic metrics and extrinsic criteria.}
\label{tab:suml_auto}
\end{figure}

 We perform a meta-evaluation using Pearson's \textit{r} and Kendall's $\tau$ to compare various intrinsic automatic metrics with our extrinsic metrics. Summary-level correlation (shown in Figure \ref{tab:suml_auto}) is shown to be much lower than system-level correlation (shown in Table \ref{tab:sysl_auto}).

\begin{figure*}[h]
\centering
\includegraphics[width=\linewidth]{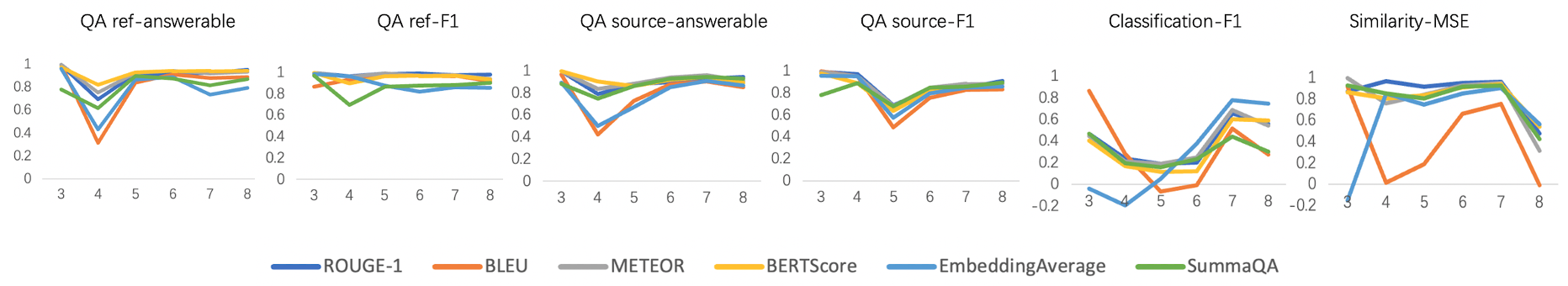}
\caption{System-level Pearson correlations between intrinsic automatic metrics and proposed extrinsic metrics on top-k systems.}
\label{tab:topk}
\end{figure*}
Our analysis reveals that there is a high correlation between extrinsic metrics in the QA task and intrinsic automatic metrics, with Pearson's \textit{r} values ranging between 0.7 and 1. Additionally, we find that there is little difference between the performance of different intrinsic automatic metrics, indicating that they are able to evaluate the QA task relatively well.

On the other hand, we observe that extrinsic metrics in classification and similarity tasks have low to moderate correlation with most intrinsic automatic metrics. The Embedding Average metric is found to be strongly correlated with the extrinsic metrics for the classification task (statistically significant at p <0.01) and show a moderate correlation for the similarity task. Other word embedding-based metrics such as Greedy Matching, Rouge-we, BERTScore and MOVERScore also show moderate correlation with extrinsic metrics in classification and similarity tasks.

In terms of the best and worst intrinsic automatic metrics, we find that no single metric consistently performs the best across all tasks. However, two intrinsic automatic metrics that are closest to the extrinsic metrics are Rouge-1 (better in the QA task) and Embedding Average (better in the similarity and classification tasks). On the other hand, CIDEr is found to be least correlated with the extrinsic metrics, and show little relevance for the similarity and classification tasks.

 We further evaluate the reliability of intrinsic automatic metrics in quantifying differences between systems with competitive performance,i.e., top-$k$ system analysis. As illustrated in Figure \ref{tab:topk}, $k$ systems are ranked based on different extrinsic metrics. We observe that for the QA-ref answerable metric and QA-source F1 and answerable metrics, the correlation between automatic and extrinsic metrics decreases slightly as the number of systems increases from 3, then increases when the number of systems reaches 5. A similar trend is also observed in the plot of the F1 indicator for the classification task, but with more noticeable fluctuations. However, we find a significant decline in the correlation between extrinsic and intrinsic automatic metrics of the similarity task as $k$ increased, which suggests that intrinsic automatic metrics should not be used to compare systems with substantial differences in usefulness in this task. While the correlation between the QA-ref answerable metric and intrinsic automatic metrics remains stable at a high level even as $k$ changed, we find that most intrinsic automatic metrics may not consistently and reliably quantify differences of usefulness between systems.

\section{Conclusions}

In this work, we conduct a user study for extrinsic evaluation of the usefulness of text summaries in different downstream tasks. Our key findings are as follows:
\begin{enumerate}
\item The usefulness of summaries is demonstrated through the dual factors of time-saving and performance. While summaries notably decrease task completion time, they may also lead to a decrease in task performance in some cases. However, the overall benefit of summaries is still apparent when considering the balance between time saved and reduction in accuracy.
\item  Summaries are particularly useful for classification and similarity tasks while being less effective for question answering tasks. This is because classification and similarity tasks rely on overall judgments of the text and do not require as much detailed information as question answering.
\item  Summaries generated by fine-tuned models exhibit consistent utility across various tasks, as they are insensitive to task differences and have a stable ranking that resembles human summaries. Conversely, zero-shot and simple extractive summaries demonstrate varying rankings across tasks.
\item  Summaries generated by fine-tuned models tend to perform better on QA tasks, while summaries generated by models in the zero-shot setting are more suitable for classification and similarity tasks. This is due to the fact that summaries generated by fine-tuned models are extractive and specific, including details such as times, places, and numbers, while summaries generated by models in the zero-shot setting are more general.
\item Intrinsic automatic metrics are suitable for assessing usefulness of summaries in QA tasks, but their utility may be limited when it comes to tasks where people are required to make an overall judgment about the text, such as classification and similarity tasks.

\end{enumerate}

\appendix
\section{Length of summaries from different systems}
\begin{figure} [b]
	\centering
        \captionsetup[subfloat]{justification=centering}
	\subfloat (a) Question answering task{
		\includegraphics[width=0.75\linewidth]{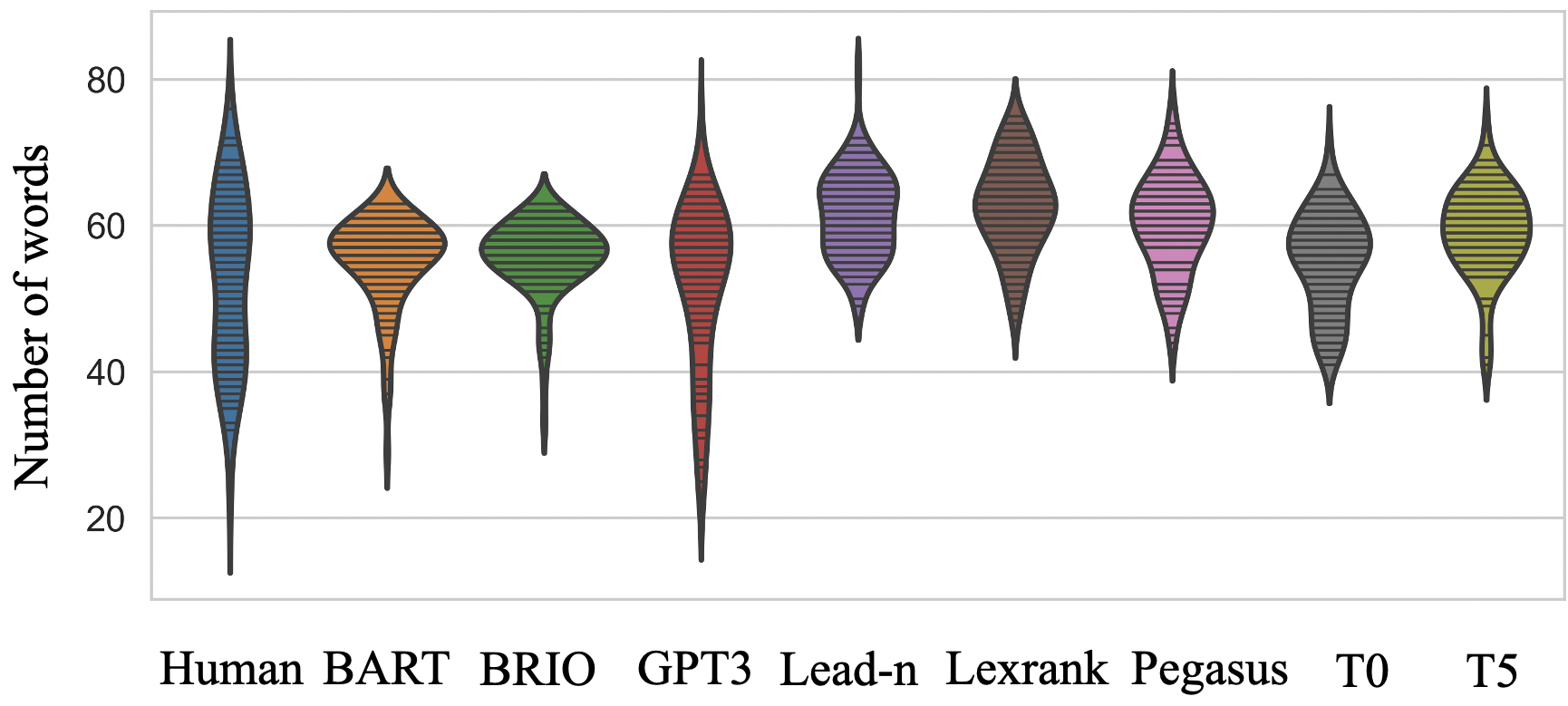}}
	\\
	\subfloat (b) Classification task{
		\includegraphics[width=0.75\linewidth]{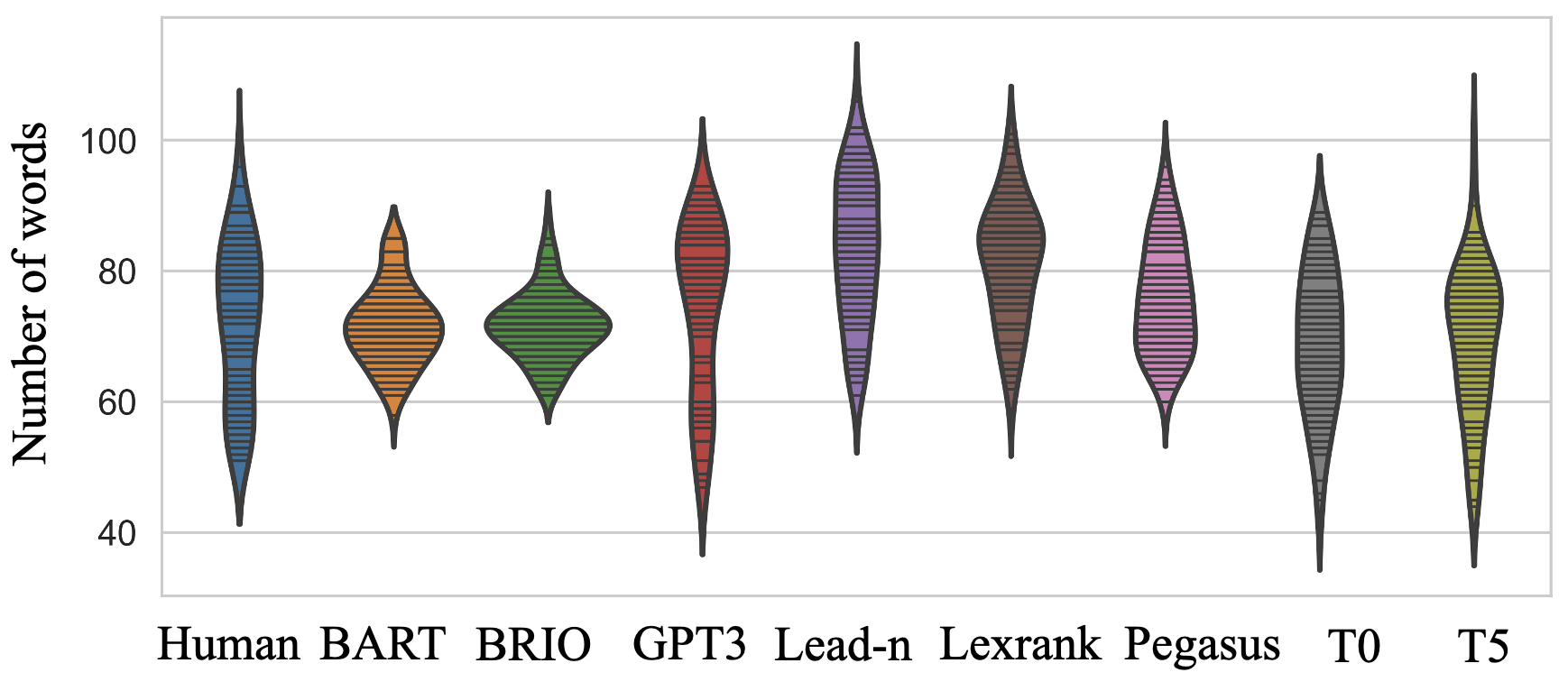} }
        \\
        \subfloat c) Similarity assessment task{
		\includegraphics[width=0.75\linewidth]{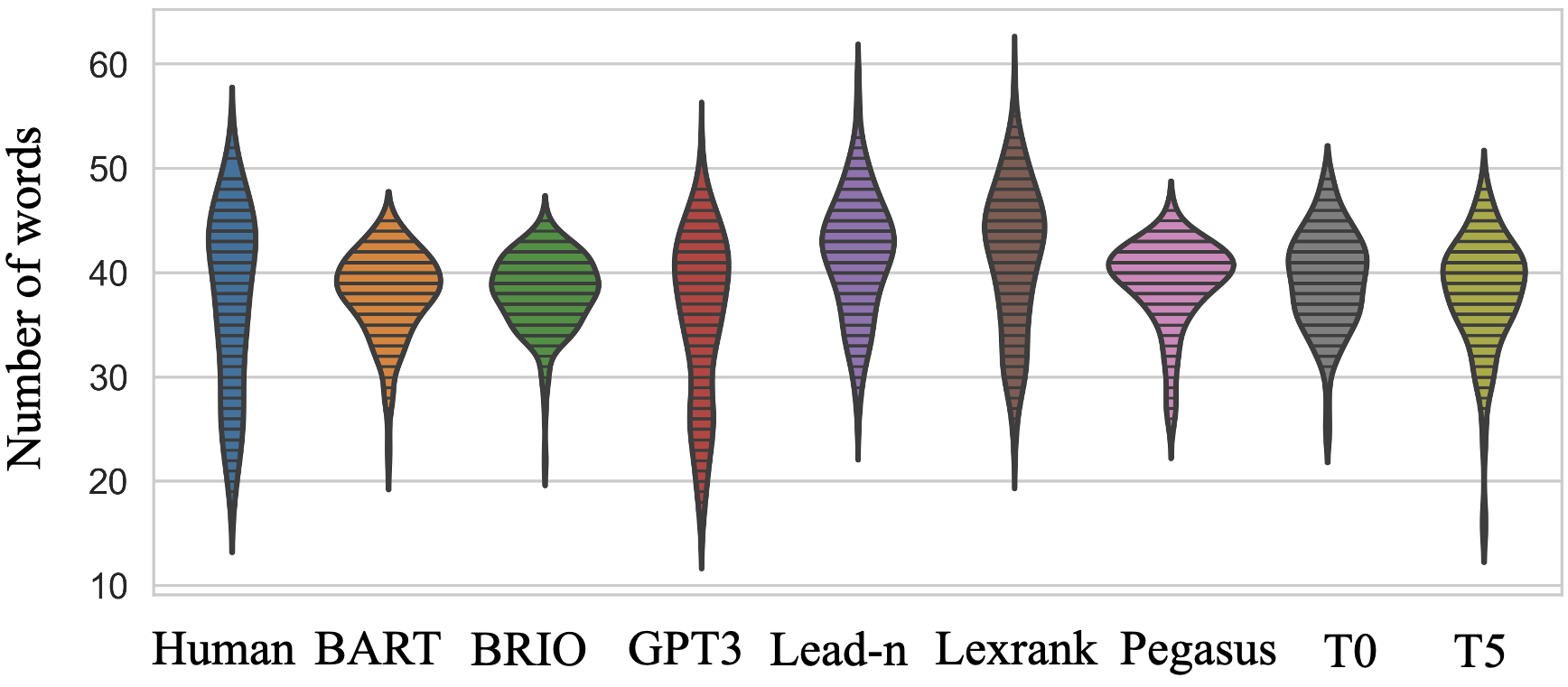} }
	\caption{Length of summaries from different systems in three tasks.}
	\label{tab:sum_len} 
\end{figure}

The length of the summary can affect the information contained in the text. Therefore, in order to ensure fairness in comparing summaries across different systems, we set a range for the number of words in the generated summary based on the length of the reference summary, so that as shown in figure \ref{tab:sum_len}, the summaries of all systems fall within a similar length interval.

\section{Correlation between extrinsic metrics}
\begin{figure}[h]
\centering
\includegraphics[width=0.7\linewidth]{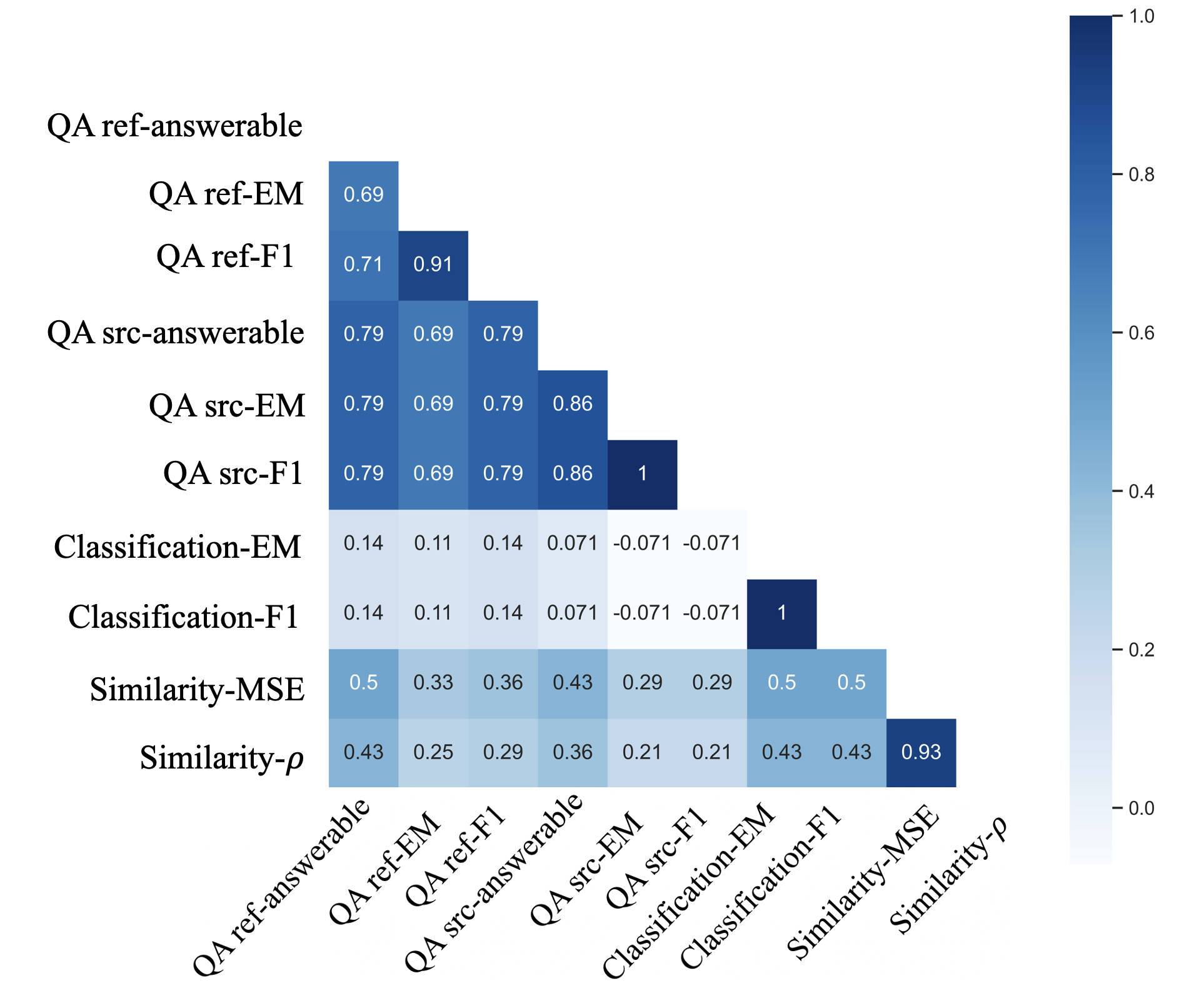}
\caption{System-level Kendall correlations of all extrinsic metrics.}
\label{tab:ex-sysl-kendall}
\end{figure}

\begin{figure}[h]
\centering
\includegraphics[width=0.44\linewidth]{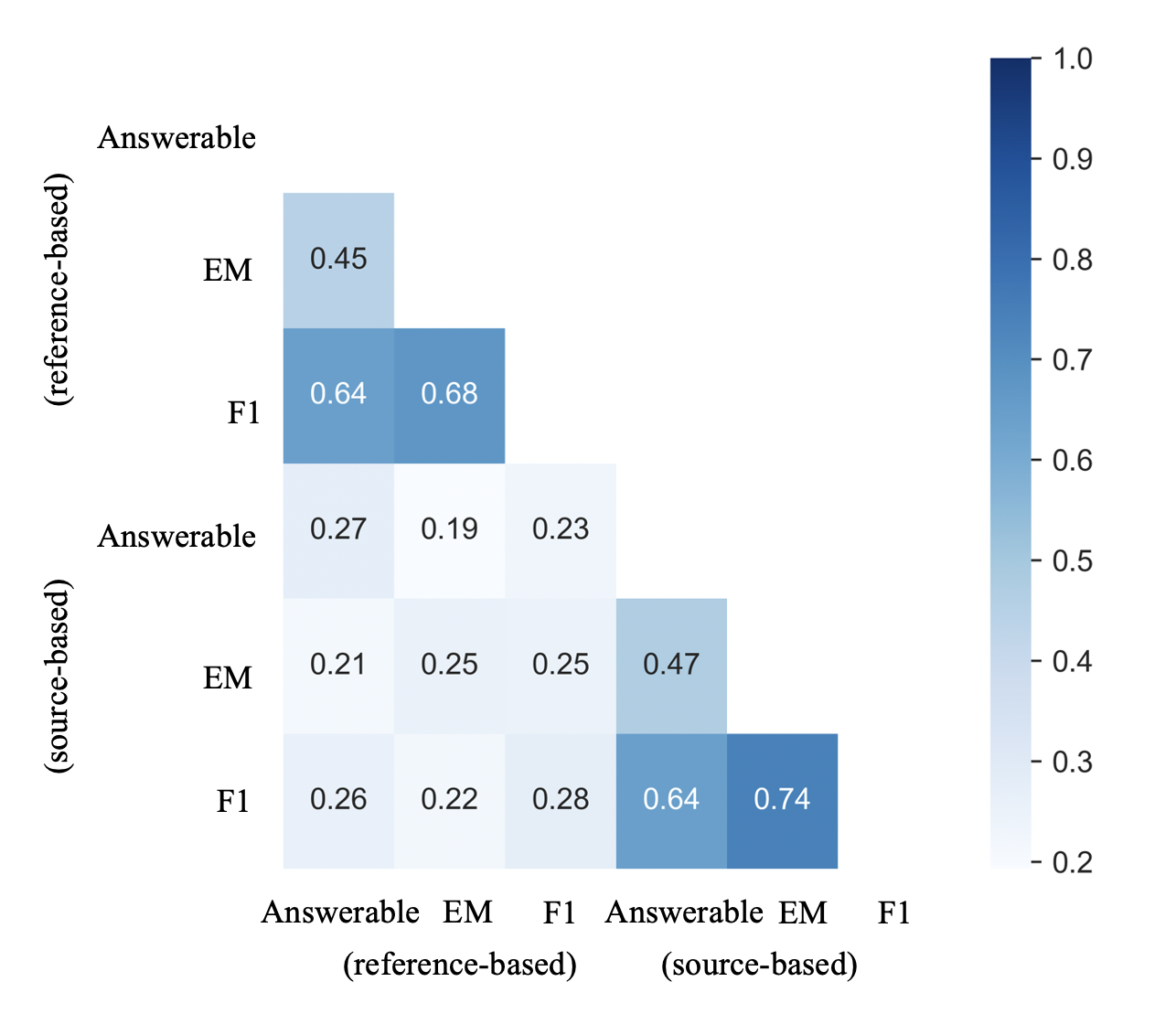}
\includegraphics[width=0.48\linewidth]{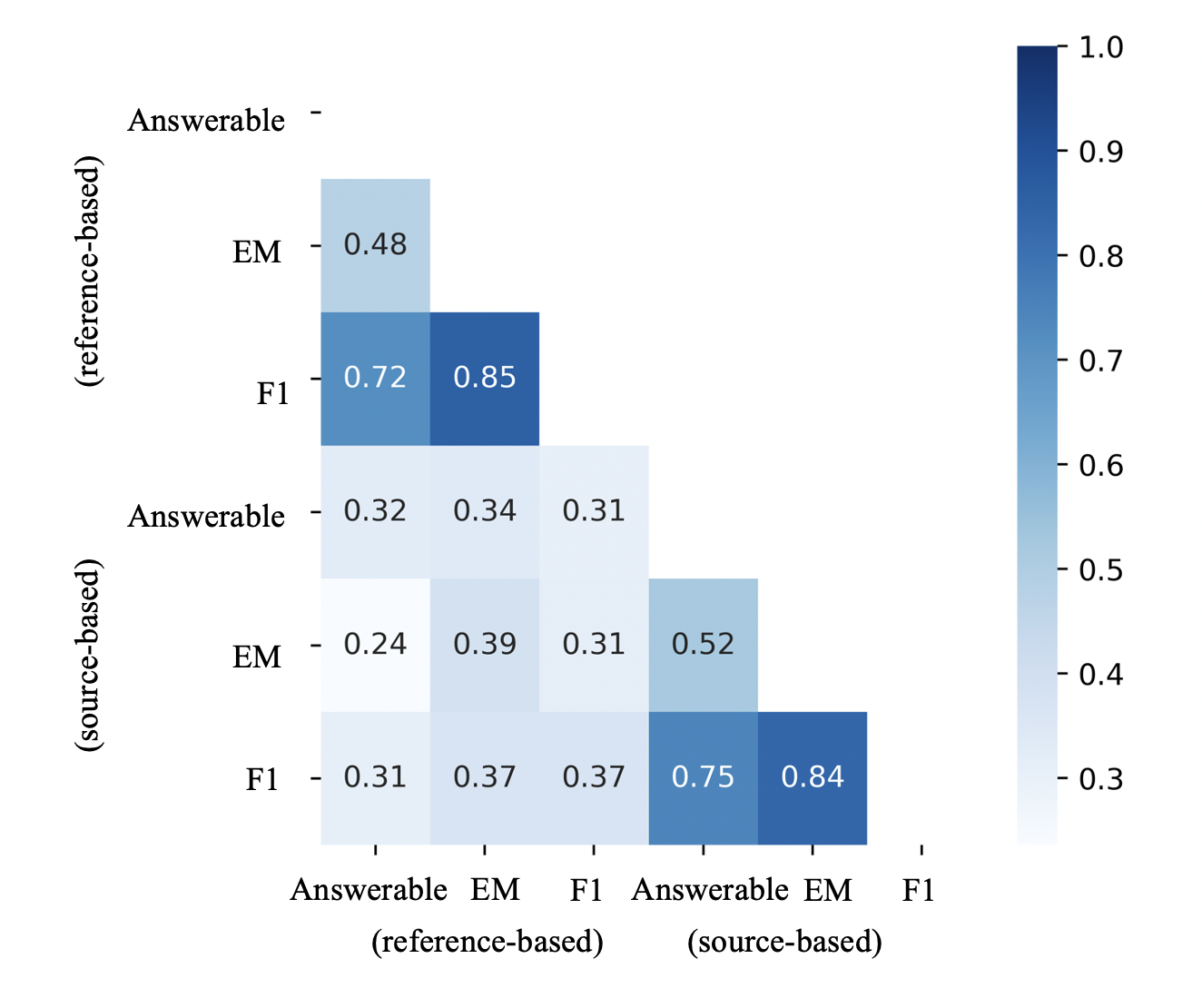}
\caption{Summary-level Kendall(left) and Pearson(right) correlations of extrinsic metrics in the QA task.}
\label{tab:ex-suml}
\end{figure}

Here we report summary-level correlations between proposed extrinsic metrics with Kendall's $\tau$ and Pearson's \textit{r} (shown in Figure \ref{tab:ex-suml}) and summary-level correlations between proposed extrinsic metrics with Pearson's \textit{r} (shown in Figure \ref{tab:ex-sysl-kendall}).

\bibliographystyle{ACM-Reference-Format}
\bibliography{sample-base}


\begin{thebibliography}{47}


\ifx \showCODEN    \undefined \def \showCODEN     #1{\unskip}     \fi
\ifx \showDOI      \undefined \def \showDOI       #1{#1}\fi
\ifx \showISBNx    \undefined \def \showISBNx     #1{\unskip}     \fi
\ifx \showISBNxiii \undefined \def \showISBNxiii  #1{\unskip}     \fi
\ifx \showISSN     \undefined \def \showISSN      #1{\unskip}     \fi
\ifx \showLCCN     \undefined \def \showLCCN      #1{\unskip}     \fi
\ifx \shownote     \undefined \def \shownote      #1{#1}          \fi
\ifx \showarticletitle \undefined \def \showarticletitle #1{#1}   \fi
\ifx \showURL      \undefined \def \showURL       {\relax}        \fi
\providecommand\bibfield[2]{#2}
\providecommand\bibinfo[2]{#2}
\providecommand\natexlab[1]{#1}
\providecommand\showeprint[2][]{arXiv:#2}

\bibitem[Banerjee and Lavie(2005)]%
        {meteor}
\bibfield{author}{\bibinfo{person}{Satanjeev Banerjee} {and}
  \bibinfo{person}{Alon Lavie}.} \bibinfo{year}{2005}\natexlab{}.
\newblock \showarticletitle{METEOR: An automatic metric for MT evaluation with
  improved correlation with human judgments}. In
  \bibinfo{booktitle}{\emph{Proceedings of the acl workshop on intrinsic and
  extrinsic evaluation measures for machine translation and/or summarization}}.
  \bibinfo{pages}{65--72}.
\newblock


\bibitem[Bhandari et~al\mbox{.}(2020)]%
        {Bhandari2020}
\bibfield{author}{\bibinfo{person}{Manik Bhandari}, \bibinfo{person}{Pranav
  Gour}, \bibinfo{person}{Atabak Ashfaq}, \bibinfo{person}{Pengfei Liu}, {and}
  \bibinfo{person}{Graham Neubig}.} \bibinfo{year}{2020}\natexlab{}.
\newblock \showarticletitle{Re-evaluating evaluation in text summarization}.
\newblock \bibinfo{journal}{\emph{arXiv preprint arXiv:2010.07100}}
  (\bibinfo{year}{2020}).
\newblock


\bibitem[Brown et~al\mbox{.}(2020)]%
        {GPT3}
\bibfield{author}{\bibinfo{person}{Tom Brown}, \bibinfo{person}{Benjamin Mann},
  \bibinfo{person}{Nick Ryder}, \bibinfo{person}{Melanie Subbiah},
  \bibinfo{person}{Jared~D Kaplan}, \bibinfo{person}{Prafulla Dhariwal},
  \bibinfo{person}{Arvind Neelakantan}, \bibinfo{person}{Pranav Shyam},
  \bibinfo{person}{Girish Sastry}, \bibinfo{person}{Amanda Askell},
  {et~al\mbox{.}}} \bibinfo{year}{2020}\natexlab{}.
\newblock \showarticletitle{Language models are few-shot learners}.
\newblock \bibinfo{journal}{\emph{Advances in neural information processing
  systems}}  \bibinfo{volume}{33} (\bibinfo{year}{2020}),
  \bibinfo{pages}{1877--1901}.
\newblock


\bibitem[Chen et~al\mbox{.}(2022)]%
        {semeval}
\bibfield{author}{\bibinfo{person}{Xi Chen}, \bibinfo{person}{Ali Zeynali},
  \bibinfo{person}{Chico Camargo}, \bibinfo{person}{Fabian Fl{\"o}ck},
  \bibinfo{person}{Devin Gaffney}, \bibinfo{person}{Przemyslaw Grabowicz},
  \bibinfo{person}{Scott Hale}, \bibinfo{person}{David Jurgens}, {and}
  \bibinfo{person}{Mattia Samory}.} \bibinfo{year}{2022}\natexlab{}.
\newblock \showarticletitle{{S}em{E}val-2022 Task 8: Multilingual news article
  similarity}. In \bibinfo{booktitle}{\emph{Proceedings of the 16th
  International Workshop on Semantic Evaluation (SemEval-2022)}}.
  \bibinfo{publisher}{Association for Computational Linguistics},
  \bibinfo{address}{Seattle, United States}, \bibinfo{pages}{1094--1106}.
\newblock
\urldef\tempurl%
\url{https://doi.org/10.18653/v1/2022.semeval-1.155}
\showDOI{\tempurl}


\bibitem[Clark et~al\mbox{.}(2019)]%
        {clark2019sentence}
\bibfield{author}{\bibinfo{person}{Elizabeth Clark}, \bibinfo{person}{Asli
  Celikyilmaz}, {and} \bibinfo{person}{Noah~A Smith}.}
  \bibinfo{year}{2019}\natexlab{}.
\newblock \showarticletitle{Sentence mover’s similarity: Automatic evaluation
  for multi-sentence texts}. In \bibinfo{booktitle}{\emph{Proceedings of the
  57th Annual Meeting of the Association for Computational Linguistics}}.
  \bibinfo{pages}{2748--2760}.
\newblock


\bibitem[Dorr et~al\mbox{.}(2005)]%
        {dorr2005methodology}
\bibfield{author}{\bibinfo{person}{Bonnie Dorr}, \bibinfo{person}{Christof
  Monz}, \bibinfo{person}{Richard Schwartz}, {and} \bibinfo{person}{David
  Zajic}.} \bibinfo{year}{2005}\natexlab{}.
\newblock \showarticletitle{A methodology for extrinsic evaluation of text
  summarization: does ROUGE correlate?}. In
  \bibinfo{booktitle}{\emph{Proceedings of the ACL Workshop on Intrinsic and
  Extrinsic Evaluation Measures for Machine Translation and/or Summarization}}.
  \bibinfo{pages}{1--8}.
\newblock


\bibitem[Erkan and Radev(2004)]%
        {Lexrank}
\bibfield{author}{\bibinfo{person}{G{\"u}nes Erkan} {and}
  \bibinfo{person}{Dragomir~R Radev}.} \bibinfo{year}{2004}\natexlab{}.
\newblock \showarticletitle{Lexrank: Graph-based lexical centrality as salience
  in text summarization}.
\newblock \bibinfo{journal}{\emph{Journal of artificial intelligence research}}
   \bibinfo{volume}{22} (\bibinfo{year}{2004}), \bibinfo{pages}{457--479}.
\newblock


\bibitem[Fabbri et~al\mbox{.}(2021)]%
        {Fabbri2021}
\bibfield{author}{\bibinfo{person}{Alexander~R Fabbri},
  \bibinfo{person}{Wojciech Kry{\'s}ci{\'n}ski}, \bibinfo{person}{Bryan
  McCann}, \bibinfo{person}{Caiming Xiong}, \bibinfo{person}{Richard Socher},
  {and} \bibinfo{person}{Dragomir Radev}.} \bibinfo{year}{2021}\natexlab{}.
\newblock \showarticletitle{Summeval: Re-evaluating summarization evaluation}.
\newblock \bibinfo{journal}{\emph{Transactions of the Association for
  Computational Linguistics}}  \bibinfo{volume}{9} (\bibinfo{year}{2021}),
  \bibinfo{pages}{391--409}.
\newblock


\bibitem[Forgues et~al\mbox{.}(2014)]%
        {forgues2014bootstrapping}
\bibfield{author}{\bibinfo{person}{Gabriel Forgues}, \bibinfo{person}{Joelle
  Pineau}, \bibinfo{person}{Jean-Marie Larchev{\^e}que}, {and}
  \bibinfo{person}{R{\'e}al Tremblay}.} \bibinfo{year}{2014}\natexlab{}.
\newblock \showarticletitle{Bootstrapping dialog systems with word embeddings}.
  In \bibinfo{booktitle}{\emph{Nips, modern machine learning and natural
  language processing workshop}}, Vol.~\bibinfo{volume}{2}.
  \bibinfo{pages}{168}.
\newblock


\bibitem[Gillick and Liu(2010)]%
        {gillick2010non}
\bibfield{author}{\bibinfo{person}{Dan Gillick} {and} \bibinfo{person}{Yang
  Liu}.} \bibinfo{year}{2010}\natexlab{}.
\newblock \showarticletitle{Non-expert evaluation of summarization systems is
  risky}. In \bibinfo{booktitle}{\emph{Proceedings of the NAACL HLT 2010
  Workshop on Creating Speech and Language Data with Amazon’s Mechanical
  Turk}}. \bibinfo{pages}{148--151}.
\newblock


\bibitem[Goyal et~al\mbox{.}(2022)]%
        {Goyal2022}
\bibfield{author}{\bibinfo{person}{Tanya Goyal}, \bibinfo{person}{Junyi~Jessy
  Li}, {and} \bibinfo{person}{Greg Durrett}.} \bibinfo{year}{2022}\natexlab{}.
\newblock \showarticletitle{News summarization and evaluation in the era of
  gpt-3}.
\newblock \bibinfo{journal}{\emph{arXiv preprint arXiv:2209.12356}}
  (\bibinfo{year}{2022}).
\newblock


\bibitem[He et~al\mbox{.}(2022)]%
        {he2022blind}
\bibfield{author}{\bibinfo{person}{Tianxing He}, \bibinfo{person}{Jingyu
  Zhang}, \bibinfo{person}{Tianle Wang}, \bibinfo{person}{Sachin Kumar},
  \bibinfo{person}{Kyunghyun Cho}, \bibinfo{person}{James Glass}, {and}
  \bibinfo{person}{Yulia Tsvetkov}.} \bibinfo{year}{2022}\natexlab{}.
\newblock \showarticletitle{On the Blind Spots of Model-Based Evaluation
  Metrics for Text Generation}.
\newblock \bibinfo{journal}{\emph{arXiv preprint arXiv:2212.10020}}
  (\bibinfo{year}{2022}).
\newblock


\bibitem[Hermann et~al\mbox{.}(2015)]%
        {Hermann2015}
\bibfield{author}{\bibinfo{person}{Karl~Moritz Hermann}, \bibinfo{person}{Tomas
  Kocisky}, \bibinfo{person}{Edward Grefenstette}, \bibinfo{person}{Lasse
  Espeholt}, \bibinfo{person}{Will Kay}, \bibinfo{person}{Mustafa Suleyman},
  {and} \bibinfo{person}{Phil Blunsom}.} \bibinfo{year}{2015}\natexlab{}.
\newblock \showarticletitle{Teaching machines to read and comprehend}.
\newblock \bibinfo{journal}{\emph{Advances in neural information processing
  systems}}  \bibinfo{volume}{28} (\bibinfo{year}{2015}).
\newblock


\bibitem[Hirao et~al\mbox{.}(2001)]%
        {hirao2001extrinsic}
\bibfield{author}{\bibinfo{person}{Tsutomu Hirao}, \bibinfo{person}{Yutaka
  Sasaki}, {and} \bibinfo{person}{Hideki Isozaki}.}
  \bibinfo{year}{2001}\natexlab{}.
\newblock \showarticletitle{An extrinsic evaluation for question-biased text
  summarization on QA tasks}. In \bibinfo{booktitle}{\emph{Proc. of the NAACL
  2001 Workshop on Automatic Summarization}}. \bibinfo{pages}{61--68}.
\newblock


\bibitem[Hovy and Lin(1998)]%
        {hovy1998}
\bibfield{author}{\bibinfo{person}{Eduard Hovy} {and} \bibinfo{person}{Chin-Yew
  Lin}.} \bibinfo{year}{1998}\natexlab{}.
\newblock \bibinfo{booktitle}{\emph{Automated text summarization and the
  SUMMARIST system}}.
\newblock \bibinfo{type}{{T}echnical {R}eport}.
  \bibinfo{institution}{UNIVERSITY OF SOUTHERN CALIFORNIA MARINA DEL REY
  INFORMATION SCIENCES INST}.
\newblock


\bibitem[Kolluru and Gotoh(2005)]%
        {kolluru2005subjectivity}
\bibfield{author}{\bibinfo{person}{BalaKrishna Kolluru} {and}
  \bibinfo{person}{Yoshihiko Gotoh}.} \bibinfo{year}{2005}\natexlab{}.
\newblock \showarticletitle{On the Subjectivity of Human Authored Summaries}.
  In \bibinfo{booktitle}{\emph{Proceedings of the ACL Workshop on Intrinsic and
  Extrinsic Evaluation Measures for Machine Translation and/or Summarization}}.
  \bibinfo{pages}{9--16}.
\newblock


\bibitem[Kryscinski et~al\mbox{.}(2020)]%
        {kryscinski-etal-2020-evaluating}
\bibfield{author}{\bibinfo{person}{Wojciech Kryscinski}, \bibinfo{person}{Bryan
  McCann}, \bibinfo{person}{Caiming Xiong}, {and} \bibinfo{person}{Richard
  Socher}.} \bibinfo{year}{2020}\natexlab{}.
\newblock \showarticletitle{Evaluating the Factual Consistency of Abstractive
  Text Summarization}. In \bibinfo{booktitle}{\emph{Proceedings of the 2020
  Conference on Empirical Methods in Natural Language Processing (EMNLP)}}.
  \bibinfo{publisher}{Association for Computational Linguistics},
  \bibinfo{address}{Online}, \bibinfo{pages}{9332--9346}.
\newblock
\urldef\tempurl%
\url{https://doi.org/10.18653/v1/2020.emnlp-main.750}
\showDOI{\tempurl}


\bibitem[Kyoomarsi et~al\mbox{.}(2008)]%
        {kyoomarsi2008optimizing}
\bibfield{author}{\bibinfo{person}{Farshad Kyoomarsi}, \bibinfo{person}{Hamid
  Khosravi}, \bibinfo{person}{Esfandiar Eslami},
  \bibinfo{person}{Pooya~Khosravyan Dehkordy}, {and} \bibinfo{person}{Asghar
  Tajoddin}.} \bibinfo{year}{2008}\natexlab{}.
\newblock \showarticletitle{Optimizing text summarization based on fuzzy
  logic}. In \bibinfo{booktitle}{\emph{Seventh IEEE/ACIS International
  Conference on Computer and Information Science (icis 2008)}}. IEEE,
  \bibinfo{pages}{347--352}.
\newblock


\bibitem[Landauer and Dumais(1997)]%
        {landauer1997solution}
\bibfield{author}{\bibinfo{person}{Thomas~K Landauer} {and}
  \bibinfo{person}{Susan~T Dumais}.} \bibinfo{year}{1997}\natexlab{}.
\newblock \showarticletitle{A solution to Plato's problem: The latent semantic
  analysis theory of acquisition, induction, and representation of knowledge.}
\newblock \bibinfo{journal}{\emph{Psychological review}} \bibinfo{volume}{104},
  \bibinfo{number}{2} (\bibinfo{year}{1997}), \bibinfo{pages}{211}.
\newblock


\bibitem[Lewis et~al\mbox{.}(2019)]%
        {BART}
\bibfield{author}{\bibinfo{person}{Mike Lewis}, \bibinfo{person}{Yinhan Liu},
  \bibinfo{person}{Naman Goyal}, \bibinfo{person}{Marjan Ghazvininejad},
  \bibinfo{person}{Abdelrahman Mohamed}, \bibinfo{person}{Omer Levy},
  \bibinfo{person}{Ves Stoyanov}, {and} \bibinfo{person}{Luke Zettlemoyer}.}
  \bibinfo{year}{2019}\natexlab{}.
\newblock \showarticletitle{Bart: Denoising sequence-to-sequence pre-training
  for natural language generation, translation, and comprehension}.
\newblock \bibinfo{journal}{\emph{arXiv preprint arXiv:1910.13461}}
  (\bibinfo{year}{2019}).
\newblock


\bibitem[Lin(2004)]%
        {lin2004rouge}
\bibfield{author}{\bibinfo{person}{Chin-Yew Lin}.}
  \bibinfo{year}{2004}\natexlab{}.
\newblock \showarticletitle{Rouge: A package for automatic evaluation of
  summaries}. In \bibinfo{booktitle}{\emph{Text summarization branches out}}.
  \bibinfo{pages}{74--81}.
\newblock


\bibitem[Liu(2019)]%
        {liu2019fine}
\bibfield{author}{\bibinfo{person}{Yang Liu}.} \bibinfo{year}{2019}\natexlab{}.
\newblock \showarticletitle{Fine-tune BERT for extractive summarization}.
\newblock \bibinfo{journal}{\emph{arXiv preprint arXiv:1903.10318}}
  (\bibinfo{year}{2019}).
\newblock


\bibitem[Liu and Lapata(2019)]%
        {liu-lapata-2019-text}
\bibfield{author}{\bibinfo{person}{Yang Liu} {and} \bibinfo{person}{Mirella
  Lapata}.} \bibinfo{year}{2019}\natexlab{}.
\newblock \showarticletitle{Text Summarization with Pretrained Encoders}. In
  \bibinfo{booktitle}{\emph{Proceedings of the 2019 Conference on Empirical
  Methods in Natural Language Processing and the 9th International Joint
  Conference on Natural Language Processing (EMNLP-IJCNLP)}}.
  \bibinfo{publisher}{Association for Computational Linguistics},
  \bibinfo{address}{Hong Kong, China}, \bibinfo{pages}{3730--3740}.
\newblock
\urldef\tempurl%
\url{https://doi.org/10.18653/v1/D19-1387}
\showDOI{\tempurl}


\bibitem[Liu et~al\mbox{.}(2022)]%
        {liu2022BRIO}
\bibfield{author}{\bibinfo{person}{Yixin Liu}, \bibinfo{person}{Pengfei Liu},
  \bibinfo{person}{Dragomir Radev}, {and} \bibinfo{person}{Graham Neubig}.}
  \bibinfo{year}{2022}\natexlab{}.
\newblock \showarticletitle{BRIO: Bringing order to abstractive summarization}.
\newblock \bibinfo{journal}{\emph{arXiv preprint arXiv:2203.16804}}
  (\bibinfo{year}{2022}).
\newblock


\bibitem[Mashechkin et~al\mbox{.}(2011)]%
        {mashechkin2011automatic}
\bibfield{author}{\bibinfo{person}{Igor~V Mashechkin}, \bibinfo{person}{MI
  Petrovskiy}, \bibinfo{person}{DS Popov}, {and} \bibinfo{person}{Dmitry~V
  Tsarev}.} \bibinfo{year}{2011}\natexlab{}.
\newblock \showarticletitle{Automatic text summarization using latent semantic
  analysis}.
\newblock \bibinfo{journal}{\emph{Programming and Computer Software}}
  \bibinfo{volume}{37} (\bibinfo{year}{2011}), \bibinfo{pages}{299--305}.
\newblock


\bibitem[Mihalcea and Tarau(2004)]%
        {mihalcea2004textrank}
\bibfield{author}{\bibinfo{person}{Rada Mihalcea} {and} \bibinfo{person}{Paul
  Tarau}.} \bibinfo{year}{2004}\natexlab{}.
\newblock \showarticletitle{Textrank: Bringing order into text}. In
  \bibinfo{booktitle}{\emph{Proceedings of the 2004 conference on empirical
  methods in natural language processing}}. \bibinfo{pages}{404--411}.
\newblock


\bibitem[Nallapati et~al\mbox{.}(2017)]%
        {nallapati2017summarunner}
\bibfield{author}{\bibinfo{person}{Ramesh Nallapati}, \bibinfo{person}{Feifei
  Zhai}, {and} \bibinfo{person}{Bowen Zhou}.} \bibinfo{year}{2017}\natexlab{}.
\newblock \showarticletitle{Summarunner: A recurrent neural network based
  sequence model for extractive summarization of documents}. In
  \bibinfo{booktitle}{\emph{Proceedings of the AAAI conference on artificial
  intelligence}}, Vol.~\bibinfo{volume}{31}.
\newblock


\bibitem[Nallapati et~al\mbox{.}(2016)]%
        {Nallapati2016}
\bibfield{author}{\bibinfo{person}{Ramesh Nallapati}, \bibinfo{person}{Bowen
  Zhou}, \bibinfo{person}{Caglar Gulcehre}, \bibinfo{person}{Bing Xiang},
  {et~al\mbox{.}}} \bibinfo{year}{2016}\natexlab{}.
\newblock \showarticletitle{Abstractive text summarization using
  sequence-to-sequence rnns and beyond}.
\newblock \bibinfo{journal}{\emph{arXiv preprint arXiv:1602.06023}}
  (\bibinfo{year}{2016}).
\newblock


\bibitem[Narayan et~al\mbox{.}(2018)]%
        {narayan2018ranking}
\bibfield{author}{\bibinfo{person}{Shashi Narayan}, \bibinfo{person}{Shay~B
  Cohen}, {and} \bibinfo{person}{Mirella Lapata}.}
  \bibinfo{year}{2018}\natexlab{}.
\newblock \showarticletitle{Ranking sentences for extractive summarization with
  reinforcement learning}.
\newblock \bibinfo{journal}{\emph{arXiv preprint arXiv:1802.08636}}
  (\bibinfo{year}{2018}).
\newblock


\bibitem[Nenkova and Passonneau(2004)]%
        {Nenkova2004}
\bibfield{author}{\bibinfo{person}{Ani Nenkova} {and}
  \bibinfo{person}{Rebecca~J Passonneau}.} \bibinfo{year}{2004}\natexlab{}.
\newblock \showarticletitle{Evaluating content selection in summarization: The
  pyramid method}. In \bibinfo{booktitle}{\emph{Proceedings of the human
  language technology conference of the north american chapter of the
  association for computational linguistics: Hlt-naacl 2004}}.
  \bibinfo{pages}{145--152}.
\newblock


\bibitem[Ng and Abrecht(2015)]%
        {rougewe}
\bibfield{author}{\bibinfo{person}{Jun-Ping Ng} {and} \bibinfo{person}{Viktoria
  Abrecht}.} \bibinfo{year}{2015}\natexlab{}.
\newblock \showarticletitle{Better summarization evaluation with word
  embeddings for ROUGE}.
\newblock \bibinfo{journal}{\emph{arXiv preprint arXiv:1508.06034}}
  (\bibinfo{year}{2015}).
\newblock


\bibitem[Ouyang et~al\mbox{.}(2022)]%
        {ouyang2022}
\bibfield{author}{\bibinfo{person}{Long Ouyang}, \bibinfo{person}{Jeff Wu},
  \bibinfo{person}{Xu Jiang}, \bibinfo{person}{Diogo Almeida},
  \bibinfo{person}{Carroll~L. Wainwright}, \bibinfo{person}{Pamela Mishkin},
  \bibinfo{person}{Chong Zhang}, \bibinfo{person}{Sandhini Agarwal},
  \bibinfo{person}{Katarina Slama}, \bibinfo{person}{Alex Ray},
  \bibinfo{person}{John Schulman}, \bibinfo{person}{Jacob Hilton},
  \bibinfo{person}{Fraser Kelton}, \bibinfo{person}{Luke Miller},
  \bibinfo{person}{Maddie Simens}, \bibinfo{person}{Amanda Askell},
  \bibinfo{person}{Peter Welinder}, \bibinfo{person}{Paul Christiano},
  \bibinfo{person}{Jan Leike}, {and} \bibinfo{person}{Ryan Lowe}.}
  \bibinfo{year}{2022}\natexlab{}.
\newblock \showarticletitle{Training language models to follow instructions
  with human feedback}.
\newblock  (\bibinfo{year}{2022}).
\newblock


\bibitem[Ozsoy et~al\mbox{.}(2011)]%
        {ozsoy2011text}
\bibfield{author}{\bibinfo{person}{Makbule~Gulcin Ozsoy},
  \bibinfo{person}{Ferda~Nur Alpaslan}, {and} \bibinfo{person}{Ilyas Cicekli}.}
  \bibinfo{year}{2011}\natexlab{}.
\newblock \showarticletitle{Text summarization using latent semantic analysis}.
\newblock \bibinfo{journal}{\emph{Journal of Information Science}}
  \bibinfo{volume}{37}, \bibinfo{number}{4} (\bibinfo{year}{2011}),
  \bibinfo{pages}{405--417}.
\newblock


\bibitem[Papineni et~al\mbox{.}(2002)]%
        {bleu}
\bibfield{author}{\bibinfo{person}{Kishore Papineni}, \bibinfo{person}{Salim
  Roukos}, \bibinfo{person}{Todd Ward}, {and} \bibinfo{person}{Wei-Jing Zhu}.}
  \bibinfo{year}{2002}\natexlab{}.
\newblock \showarticletitle{Bleu: a method for automatic evaluation of machine
  translation}. In \bibinfo{booktitle}{\emph{Proceedings of the 40th annual
  meeting of the Association for Computational Linguistics}}.
  \bibinfo{pages}{311--318}.
\newblock


\bibitem[Popovi{\'c}(2017)]%
        {chrf}
\bibfield{author}{\bibinfo{person}{Maja Popovi{\'c}}.}
  \bibinfo{year}{2017}\natexlab{}.
\newblock \showarticletitle{chr{F}++: words helping character n-grams}. In
  \bibinfo{booktitle}{\emph{Proceedings of the Second Conference on Machine
  Translation}}. \bibinfo{publisher}{Association for Computational
  Linguistics}, \bibinfo{address}{Copenhagen, Denmark},
  \bibinfo{pages}{612--618}.
\newblock
\urldef\tempurl%
\url{https://doi.org/10.18653/v1/W17-4770}
\showDOI{\tempurl}


\bibitem[Raffel et~al\mbox{.}(2020)]%
        {T5}
\bibfield{author}{\bibinfo{person}{Colin Raffel}, \bibinfo{person}{Noam
  Shazeer}, \bibinfo{person}{Adam Roberts}, \bibinfo{person}{Katherine Lee},
  \bibinfo{person}{Sharan Narang}, \bibinfo{person}{Michael Matena},
  \bibinfo{person}{Yanqi Zhou}, \bibinfo{person}{Wei Li}, {and}
  \bibinfo{person}{Peter~J Liu}.} \bibinfo{year}{2020}\natexlab{}.
\newblock \showarticletitle{Exploring the limits of transfer learning with a
  unified text-to-text transformer}.
\newblock \bibinfo{journal}{\emph{The Journal of Machine Learning Research}}
  \bibinfo{volume}{21}, \bibinfo{number}{1} (\bibinfo{year}{2020}),
  \bibinfo{pages}{5485--5551}.
\newblock


\bibitem[Rus and Lintean(2012)]%
        {greedymatching}
\bibfield{author}{\bibinfo{person}{Vasile Rus} {and} \bibinfo{person}{Mihai
  Lintean}.} \bibinfo{year}{2012}\natexlab{}.
\newblock \showarticletitle{A Comparison of Greedy and Optimal Assessment of
  Natural Language Student Input Using Word-to-Word Similarity Metrics}. In
  \bibinfo{booktitle}{\emph{Proceedings of the Seventh Workshop on Building
  Educational Applications Using {NLP}}}. \bibinfo{publisher}{Association for
  Computational Linguistics}, \bibinfo{address}{Montr{\'e}al, Canada},
  \bibinfo{pages}{157--162}.
\newblock
\urldef\tempurl%
\url{https://aclanthology.org/W12-2018}
\showURL{%
\tempurl}


\bibitem[Sandhaus(2008)]%
        {NYT}
\bibfield{author}{\bibinfo{person}{Evan Sandhaus}.}
  \bibinfo{year}{2008}\natexlab{}.
\newblock \showarticletitle{The New York Times Annotated Corpus}.
\newblock  (\bibinfo{year}{2008}).
\newblock
\urldef\tempurl%
\url{https://doi.org/10.35111/77ba-9x74}
\showDOI{\tempurl}


\bibitem[Sanh et~al\mbox{.}(2021)]%
        {T0}
\bibfield{author}{\bibinfo{person}{Victor Sanh}, \bibinfo{person}{Albert
  Webson}, \bibinfo{person}{Colin Raffel}, \bibinfo{person}{Stephen~H Bach},
  \bibinfo{person}{Lintang Sutawika}, \bibinfo{person}{Zaid Alyafeai},
  \bibinfo{person}{Antoine Chaffin}, \bibinfo{person}{Arnaud Stiegler},
  \bibinfo{person}{Teven~Le Scao}, \bibinfo{person}{Arun Raja},
  {et~al\mbox{.}}} \bibinfo{year}{2021}\natexlab{}.
\newblock \showarticletitle{Multitask prompted training enables zero-shot task
  generalization}.
\newblock \bibinfo{journal}{\emph{arXiv preprint arXiv:2110.08207}}
  (\bibinfo{year}{2021}).
\newblock


\bibitem[Scialom et~al\mbox{.}(2019)]%
        {summaqa}
\bibfield{author}{\bibinfo{person}{Thomas Scialom}, \bibinfo{person}{Sylvain
  Lamprier}, \bibinfo{person}{Benjamin Piwowarski}, {and}
  \bibinfo{person}{Jacopo Staiano}.} \bibinfo{year}{2019}\natexlab{}.
\newblock \showarticletitle{Answers unite! unsupervised metrics for reinforced
  summarization models}.
\newblock \bibinfo{journal}{\emph{arXiv preprint arXiv:1909.01610}}
  (\bibinfo{year}{2019}).
\newblock


\bibitem[Shapira et~al\mbox{.}(2019)]%
        {shapira2019crowdsourcing}
\bibfield{author}{\bibinfo{person}{Ori Shapira}, \bibinfo{person}{David Gabay},
  \bibinfo{person}{Yang Gao}, \bibinfo{person}{Hadar Ronen},
  \bibinfo{person}{Ramakanth Pasunuru}, \bibinfo{person}{Mohit Bansal},
  \bibinfo{person}{Yael Amsterdamer}, {and} \bibinfo{person}{Ido Dagan}.}
  \bibinfo{year}{2019}\natexlab{}.
\newblock \showarticletitle{Crowdsourcing lightweight pyramids for manual
  summary evaluation}.
\newblock \bibinfo{journal}{\emph{arXiv preprint arXiv:1904.05929}}
  (\bibinfo{year}{2019}).
\newblock


\bibitem[Suanmali et~al\mbox{.}(2009)]%
        {suanmali2009sentence}
\bibfield{author}{\bibinfo{person}{Ladda Suanmali},
  \bibinfo{person}{Mohammed~Salem Binwahlan}, {and} \bibinfo{person}{Naomie
  Salim}.} \bibinfo{year}{2009}\natexlab{}.
\newblock \showarticletitle{Sentence features fusion for text summarization
  using fuzzy logic}. In \bibinfo{booktitle}{\emph{2009 Ninth International
  Conference on Hybrid Intelligent Systems}}, Vol.~\bibinfo{volume}{1}. IEEE,
  \bibinfo{pages}{142--146}.
\newblock


\bibitem[Vedantam et~al\mbox{.}(2015)]%
        {cider}
\bibfield{author}{\bibinfo{person}{Ramakrishna Vedantam}, \bibinfo{person}{C
  Lawrence~Zitnick}, {and} \bibinfo{person}{Devi Parikh}.}
  \bibinfo{year}{2015}\natexlab{}.
\newblock \showarticletitle{Cider: Consensus-based image description
  evaluation}. In \bibinfo{booktitle}{\emph{Proceedings of the IEEE conference
  on computer vision and pattern recognition}}. \bibinfo{pages}{4566--4575}.
\newblock


\bibitem[Verma and Nidhi(2017)]%
        {verma2017extractive}
\bibfield{author}{\bibinfo{person}{Sukriti Verma} {and}
  \bibinfo{person}{Vagisha Nidhi}.} \bibinfo{year}{2017}\natexlab{}.
\newblock \showarticletitle{Extractive summarization using deep learning}.
\newblock \bibinfo{journal}{\emph{arXiv preprint arXiv:1708.04439}}
  (\bibinfo{year}{2017}).
\newblock


\bibitem[Zhang et~al\mbox{.}(2020)]%
        {zhang2020pegasus}
\bibfield{author}{\bibinfo{person}{Jingqing Zhang}, \bibinfo{person}{Yao Zhao},
  \bibinfo{person}{Mohammad Saleh}, {and} \bibinfo{person}{Peter Liu}.}
  \bibinfo{year}{2020}\natexlab{}.
\newblock \showarticletitle{Pegasus: Pre-training with extracted gap-sentences
  for abstractive summarization}. In \bibinfo{booktitle}{\emph{International
  Conference on Machine Learning}}. PMLR, \bibinfo{pages}{11328--11339}.
\newblock


\bibitem[Zhang et~al\mbox{.}(2019)]%
        {zhang2019bertscore}
\bibfield{author}{\bibinfo{person}{Tianyi Zhang}, \bibinfo{person}{Varsha
  Kishore}, \bibinfo{person}{Felix Wu}, \bibinfo{person}{Kilian~Q Weinberger},
  {and} \bibinfo{person}{Yoav Artzi}.} \bibinfo{year}{2019}\natexlab{}.
\newblock \showarticletitle{Bertscore: Evaluating text generation with bert}.
\newblock \bibinfo{journal}{\emph{arXiv preprint arXiv:1904.09675}}
  (\bibinfo{year}{2019}).
\newblock


\bibitem[Zhao et~al\mbox{.}(2019)]%
        {moverscore}
\bibfield{author}{\bibinfo{person}{Wei Zhao}, \bibinfo{person}{Maxime Peyrard},
  \bibinfo{person}{Fei Liu}, \bibinfo{person}{Yang Gao},
  \bibinfo{person}{Christian~M Meyer}, {and} \bibinfo{person}{Steffen Eger}.}
  \bibinfo{year}{2019}\natexlab{}.
\newblock \showarticletitle{MoverScore: Text generation evaluating with
  contextualized embeddings and earth mover distance}.
\newblock \bibinfo{journal}{\emph{arXiv preprint arXiv:1909.02622}}
  (\bibinfo{year}{2019}).
\newblock


\end{thebibliography}

\end{document}